\documentclass[runningheads]{llncs}

\usepackage{eccv}

\usepackage{eccvabbrv}

\usepackage{graphicx}
\usepackage{booktabs}

\usepackage{capt-of} 
\usepackage{ragged2e} 
\usepackage{tabularx}
\usepackage{makecell}
\usepackage{amsmath}
\usepackage[dvipsnames]{xcolor}
\usepackage{array}
\usepackage{pifont}
\usepackage{amssymb}

\usepackage[accsupp]{axessibility}  

\usepackage{hyperref}

\usepackage{orcidlink}

\begin{document}

\title{\textsc{InStyle}: Instant Appearance Stylization of 3D Shapes} 

\titlerunning{\textsc{InStyle}: Instant Appearance Stylization of 3D Shapes}

\author{Başak Melis Öcal\inst{1}\thanks{Correspondence to <\email{b.m.ocal@uva.nl}>} \and
Xiaoyan Xing\inst{1} \and 
Yue Li\inst{1} \and 
Ngo Anh Vien\inst{2} \and 
Sezer Karaoğlu\inst{1} \and 
Theo Gevers\inst{1}}

\authorrunning{B.M. Öcal et al.}

\institute{UvA-Bosch Delta Lab, University of Amsterdam \and
Bosch Center for AI, Robert Bosch GmbH
}
\newcommand{\MethodName}{{\textsc{InStyle}}}

\maketitle

\begin{figure*}[h]
   \centering
   \includegraphics[height=8.4cm]{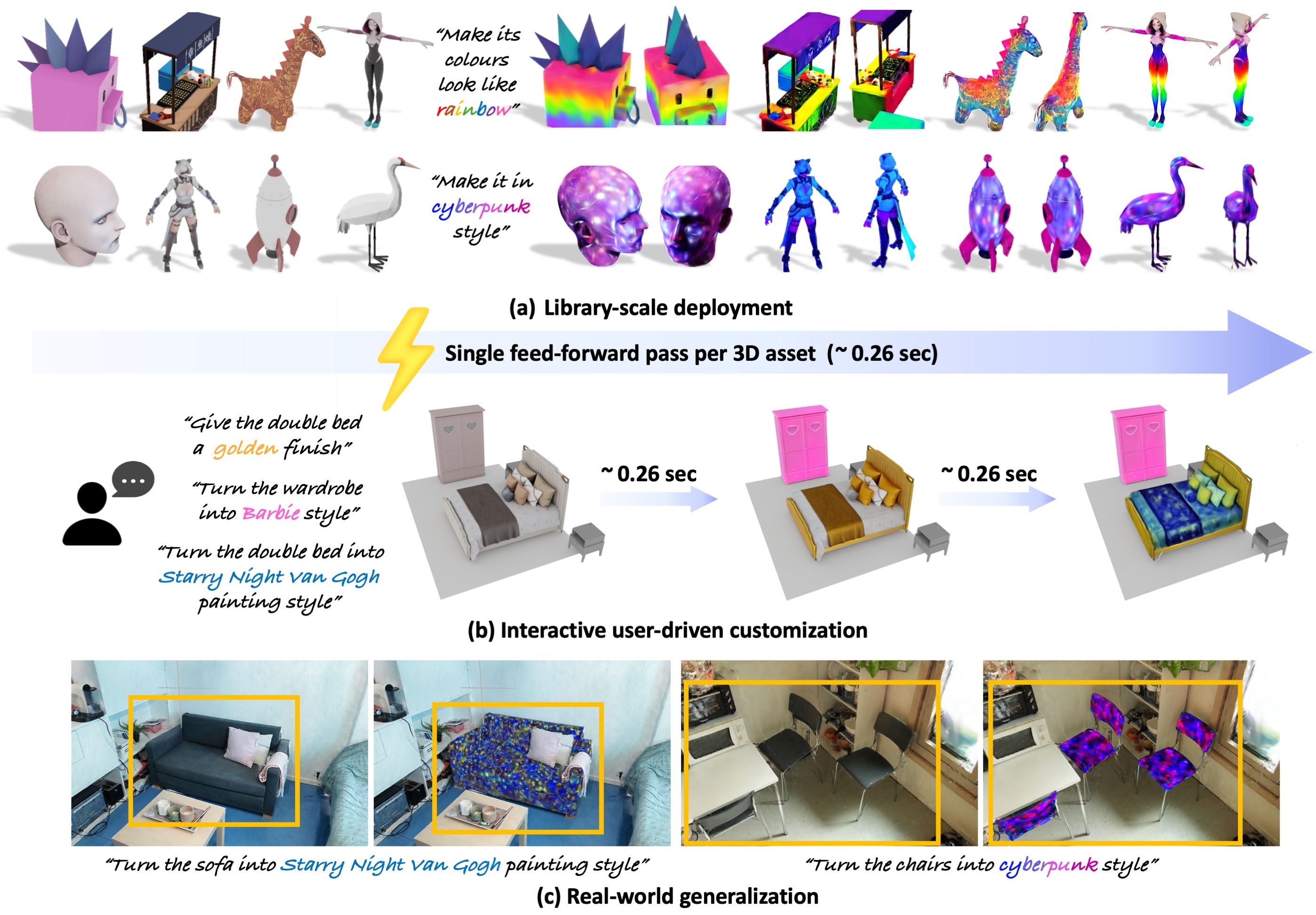}
   \captionof{figure}{Given a 3D Gaussian splat asset and an edit prompt, \MethodName{} - a diffusion-based feed-forward style editor - delivers geometry-preserving, multi-view-consistent appearance stylization with a single-pass inference (0.26\,s per asset), without per-asset test-time optimization. \MethodName{} enables (a) library-scale stylization across diverse assets and (b) interactive, user-driven customization, (c) while generalizing to real-world captures - unlocking practical, democratized 3D stylization at scale.}
   \label{fig:teaser}
   \vspace{-2.2em}
\end{figure*}

\vspace{-0.9em}
\begin{abstract}

3D appearance stylization is central to game development, AR/VR, and interior design, where large asset libraries demand scalable, fast, high-fidelity style variation while keeping geometry intact. Prior work on text-to-3D stylization distills from 2D editors, either via per-asset test-time optimization or by learning feed-forward predictors. The former is slow and often brittle, limiting interactive and library-scale use, while the latter drifts geometry during stylization. In contrast, our key insight is that amortized, feed-forward 3DGS stylization becomes feasible only in a geometry–appearance factorized latent space, where geometry acts as an identity anchor for coherent appearance edits. Building on this, we introduce \MethodName{}, a feed-forward framework for text-driven 3DGS stylization with single-pass inference. \MethodName{} learns disentangled geometry and appearance latents from spatially grouped Gaussians with controlled cross-branch sharing, and edits appearance with a text-conditioned latent diffusion editor guided by geometry. While promising, 2D-to-3D distillation is non-trivial; we therefore build on a multi-view, distribution-matching distillation objective to supervise a coherent 3D edit. Comprehensive evaluations show that \MethodName{} delivers geometry-preserving, multi-view consistent stylization and outperforms per-asset optimization methods, while generalizing to out-of-domain assets, including real-world captures - unlocking practical, interactive and library-scale stylization with sub-second inference costs per asset. Our project page is available at \href{https://instyle3d.github.io/}{https://instyle3d.github.io/}.

\end{abstract}

\section{Introduction}
\label{sec:intro}

3D style editing is a core requirement in gaming, AR/VR, interior design, and digital arts, where immersive experiences rely on large libraries of customizable assets. In many real workflows, geometry is already available (scans/CAD models) and must be preserved; so customization focuses on appearance editing, such as materials, textures, and colors, while keeping geometric identity intact. Workflows increasingly require instant editing: 1) offline to rapidly explore and batch-apply a catalog of styles during content creation and 2) online for interactive, user-driven customization. At library scale \cite{unityassetstore, sketchup3dwarehouse}, such stylization is typically specified through predefined, art-directed style catalogs to maintain consistent art direction. A common example arises in retail and interior design, where CAD catalogs evolve rapidly; manually authoring style variants for each newly added asset is laborious and expert-driven, often taking minutes to hours. These demands call for fast, multi-view-consistent, geometry-preserving stylization that scales with growing asset libraries and supports interactive use.

Recent breakthroughs in text-to-image diffusion \cite{ddpm, stablediffusion} have made 2D editing highly effective \cite{Instructp2p}, inspiring 3D extensions. A straightforward strategy is to edit rendered views with a 2D editor and then reconstruct a 3D representation. However, per-view edits often break multi-view consistency, making reconstruction ill-conditioned or drift-prone, and the render–edit–reconstruct loop remains expensive to repeat per-asset. Most 3D methods therefore follow two dominant paradigms: iterative dataset update, where edited renders supervise repeated re-optimization \cite{in2n}, and score-distillation approaches that directly optimize 3D representations with diffusion guidance \cite{dreameditor, focaldreamer, customizenerf}. While effective, both are bottlenecked by slow and often brittle per-asset optimization, making them difficult to scale to large 3D libraries and interactive use. Recent zero-shot or feed-forward alternatives \cite{freeditor, shapeditor} reduce editing time, but they typically operate on a shared geometry–appearance representation; yet controlling appearance without geometry drift remains challenging.

To address these shortcomings - slow per-scene optimization and limited identity-preserving control - we introduce \MethodName{}, a diffusion-based feed-forward 3D Gaussian splat (3DGS) style editor that operates in the latent space of a 3D Gaussian VAE. Once trained, it produces geometry-preserving edits in a single forward pass, amortizing editing cost across asset libraries and eliminating per-asset test-time optimization. Our key insight is that feed-forward, controllable 3DGS stylization becomes feasible only in a structured, geometry–appearance factorized latent space, using geometry as an identity anchor to enforce coherent appearance edits. More concretely: (1) 3DGS is an unordered set of primitives, making structure-aware latent editing challenging; we therefore group Gaussians by spatial proximity to form 3D structure-aware tokens for diffusion prior learning. (2) The geometry and appearance parameters of Gaussians follow different distributions, creating mismatched learning dynamics that make joint optimization non-trivial. (3) 3D appearance stylization should preserve shape while modifying materials, textures, and colors, yet still exploit geometric context. Leveraging these insights, we learn disentangled appearance and geometry latents via a dual-branch 3D VAE, coupled through a cross-branch feature-sharing module that exchanges only necessary signals, simplifying parameter learning and enabling controllable edits. On this latent space, a diffusion model is pre-trained over the appearance latent under a reconstruction objective, it is then adapted for prompt-driven editing via multi-view, distribution-matching distillation from a frozen 2D editor, while shape is preserved through the geometry latent.

\MethodName{} delivers high-quality, geometry-preserving, multi-view-consistent stylizations across a diverse set of 3D assets by distilling a pre-trained 2D editor into a 3D-native model. As illustrated in \cref{fig:teaser}, \MethodName{} offers single-pass, sub-second stylization that scales to large asset libraries, supports interactive user-driven customization, and handles real-world captures and complex scenes without per-asset test-time optimization. Extensive evaluations and user studies show that \MethodName{} outperforms prior methods on prompt alignment, structural consistency, and user preference. In brief, our contributions are:

\begin{itemize}
\item Feed-forward framework: A feed-forward 3DGS editor for geometry-preserving stylization that eliminates per-asset test-time optimization and runs in a single forward pass, enabling interactive use and library-scale deployment (\cref{subsec:method1}). 
\item Disentangled latent representation and learning formulation: A dual-branch 3D VAE that factorizes geometry and appearance with controlled cross-branch exchange, enabling geometry-preserving targeted edits, and stabilizing learning of Gaussian parameters. A multi-view, distribution matching distillation setup that transfers pre-trained 2D edit behavior into a single coherent 3D edit in latent space (\cref{subsec:method2}, \cref{subsec:method3}). 
\item Generalizable stylization: \MethodName{} successfully stylizes unseen 3DGS assets, including real-world captures with diverse geometry (\cref{fig:teaser}, \cref{sec:experiments}). 
\end{itemize}

\section{Related Work}
\label{relatedwork}

\subsection{Image Editing with Diffusion Models}
\label{subsec:imagediting}

Recent advances in image generation models \cite{ddpm, stablediffusion} have paved the way for diffusion-based 2D image editing with direct guidance. Personalization methods \cite{dreambooth, multiconceptcustom, animage} 
preserve identity while controlling appearance/pose, drag-based~\cite{dragfusion,mdragondiffusion} and layout-guided models~\cite{layout1,layout2,layout3} enforce spatial control. GLIDE \cite{glide} guides image generation using CLIP features, and ControlNet \cite{controlnet} incorporates diverse conditioning signals such as pose or normal maps. More relevant to our work, several studies \cite{text2live, imagic, p2p, zeroshoti2i, nulltext, sdedit, plugandplay} have reframed image editing as an image-to-image translation task guided by text or image conditions. Building on Stable Diffusion~\cite{stablediffusion}, InstructPix2Pix and variants~\cite{ip2p,hive,watchyoursteps, multireward} improve edit quality and localization via joint image–text conditioning.

\subsection{3D Generation}
\label{subsec:3dgeneration}
Inspired by text-to-image diffusion, DreamFusion~\cite{dreamfusion} first leveraged pretrained diffusion via SDS to optimize 3D content. Follow-up work improved fidelity~\cite{magic3d,fantasia3d,sweetdreamer} and paired 3DGS with 2D priors~\cite{gaussiandreamer,dreamgaussian,gaussiansplatting} to accelerate optimization. Despite progress, SDS approaches still require per-asset optimization and are prone to geometric ambiguity. Recent methods fine-tune 2D diffusion for multi-view image generation~\cite{zero123,mvdream,mvdiffusion,syncdreamer,wonder3d}, followed by 3D reconstruction. Two-stage pipelines first generate consistent views and then reconstruct 3D geometry ~\cite{instant3d, lgm}, and others diffuse over encoded multi-view splat images ~\cite{diffsplat}. However, many such pipelines rely on multi-view intermediates rather than a unified 3D latent representation, limiting 3D-consistency. Recent approaches like GVGEN \cite{gvgen} adapt diffusion to 3D Gaussians via volumetric structures, but introduce costly preprocessing and training. To improve scalability, several methods operate in 3D latent spaces~\cite{shape2vecset,clay,3dtopia,gpld3d,atlas,trellis}. Still, directly modeling 3D Gaussians remains challenging due to their unstructured, discrete nature. \MethodName{} addresses this by proximity-based grouping of Gaussians into structure-aware tokens, encoded in a geometry–appearance factorized latent space that preserves global 3D context and enables diffusion prior learning directly over 3DGS.

\subsection{Text-driven 3D Editing and Stylization}
\label{subsec:3dediting}

Recent 3D editing methods start from existing representations (\eg, NeRF or 3DGS) and target either geometry, appearance, or both, typically distilling from 2D image editors or vision-language models. We focus on geometry-preserving appearance stylization (materials, textures, color), a practically common asset-workflow setting where preventing geometry drift and ensuring multi-view consistency are key challenges. CLIP-based approaches \cite{clip,text2mesh,tango,clipnerf,nerfart,temo} align appearance or geometry with text or reference images. Built on InstructPix2Pix~\cite{ip2p}, IN2N \cite{in2n} pioneered the \textit{iterative dataset update} paradigm: edited renders supervise iterative optimization of the 3D representation. This strategy has been later extended to 3DGS \cite{igs2gs, gaussianeditor} and point clouds \cite{Instructp2p}. Subsequent research \cite{consistentdreamer, vicanerf, viewconsistent3d, editsplat} improve multi-view consistency through attention-based latent code alignment \cite{gaussctrl}, feature injection with epipolar constraints \cite{dge}, depth-guided feature-sharing \cite{morpheus}. Orthogonally, SDS-based methods directly optimize 3D~\cite{instruct3d23d,ednerf,voxe,dreameditor,focaldreamer,customizenerf}. Despite progress, iterative dataset update and score distillation approaches continue to suffer from slow per-asset optimization, limiting scalability and interactive use. ProteusNeRF \cite{proteusnerf} and Edit-DiffNeRF \cite{editdiffnerf} accelerate optimization, while FreeEditor \cite{freeditor} offers a zero-shot alternative. More related to our work, Shap-Editor \cite{shapeditor} learns a feed-forward network eliminating test-time optimization, but edits a shared appearance-geometry representation, which often couples appearance changes with geometry drift and over-smoothed textures. In contrast, \MethodName{} targets geometry-preserving stylization by disentangling geometry and appearance latents and enabling controlled cross-branch exchange, yielding view-consistent edits in a single feed-forward pass.

\section{Method}
\label{method}
\vspace{-2.9em}
\begin{figure*}[h]
  \centering
   \includegraphics[height=4.5cm]{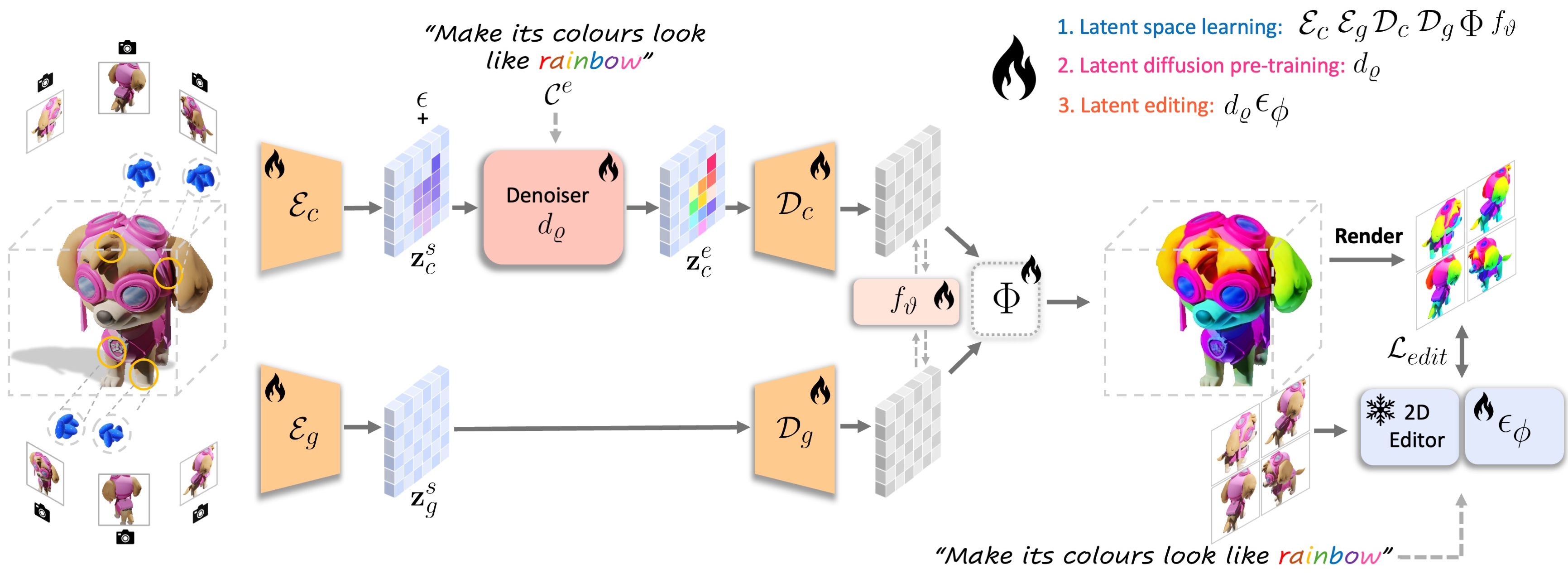}
   \caption{\textbf{Overview of our method.} \textcolor{NavyBlue}{(1) Latent space learning:} Given input Gaussians, our method first groups them based on spatial proximity and encodes into group-structured disentangled latent spaces, with controlled cross-branch feature-sharing. \textcolor{RubineRed}{(2) Latent diffusion pre-training:} A denoiser $d_{\varrho}$ then learns to denoise the noisy appearance latent ${\mathbf{z}_{c}^{s}}_\mathbf{t}$ conditioned on text embedding $\mathcal{C}$. \textcolor{RedOrange}{(3) Latent editing:} Once 3D priors are captured, $d_{\varrho}$ is further trained to learn an editing function $g(\cdot)$ that maps latent $\mathbf{z}_{c}^{s}$ to a modified latent $\mathbf{z}_{c}^{e}$ based on text embedding $\mathcal{C}^{e}$, guided by the geometry latent ${\mathbf{z}_{g}^{s}}$. At \textcolor{Emerald}{inference}, \MethodName{} generates modified high-quality, 3D-consistent assets from text prompts in a single feed-forward pass instantly, fully eliminating test-time optimization. Trainable models at each stage are denoted.}
   \label{fig:arch}
   \vspace{-1.5em}
\end{figure*}

We introduce \MethodName{}, a diffusion-based, feed-forward 3D Gaussian splat style editor that learns 3D priors directly from spatially-grouped Gaussians, encoded into disentangled latent spaces for appearance and geometry via a 3D Gaussian VAE. These priors are then used for learning an editing function $g(\cdot)$. The training pipeline is divided into three stages:
\begin{enumerate}
\item \textcolor{NavyBlue}{Latent space learning:} Dual-encoders ${\mathcal{E}_{g}}$ and $ {\mathcal{E}_{c}}$ map the spatially-grouped input Gaussians $\mathcal{G}$, into latents $\mathbf{z}_{g}^{s}$ and $\mathbf{z}_{c}^{s}$. Dual-decoders ${\mathcal{D}_{g}}$ and ${\mathcal{D}_{c}}$, together with a cross-branch feature-sharing module $f_{\vartheta}$, then reconstruct the Gaussian parameters (\cref{subsec:method1}). 
\item \textcolor{RubineRed}{Latent diffusion pre-training:} A latent diffusion model $d_{\varrho}$ learns denoising the noisy latent ${\mathbf{z}_{c}^{s}}_\mathbf{t}$ conditioned on text embedding $\mathcal{C}$, with the objective of reconstructing the input Gaussian representation $\mathcal{G}$ (\cref{subsec:method2}).
\item \textcolor{RedOrange}{Latent editing:} Once 3D priors are captured, $d_{\varrho}$ is further trained to learn an editing function $g(\cdot)$ that maps latent $\mathbf{z}_{c}^{s}$ to a modified latent $\mathbf{z}_{c}^{e}$ conditioned on text embedding $\mathcal{C}^{e}$. Supervision is provided by distilling from a frozen 2D image editor (\cref{subsec:method3}).
\end{enumerate}

\noindent The diffusion-related stages are guided by the geometry latent ${\mathbf{z}_{g}^{s}}$ through cross-branch feature-sharing module $f_{\vartheta}$ to provide structural context. At \textcolor{Emerald}{inference}, \MethodName{} generates modified 3D-consistent assets from text prompts in a single feed-forward pass instantly, fully eliminating test-time optimization. The scheme of the framework is illustrated in \cref{fig:arch}.

\subsection{Stage 1: Gaussian VAE for Group-structured Disentangled Latent Space Learning}
\label{subsec:method1}

Learning latent representations for 3D Gaussians poses two key challenges: (1) 3DGS is unstructured, hindering recovery of spatial structure, and (2) geometry and appearance parameters exhibit different distributions \cite{shapesplat}, making naive joint learning unstable. \MethodName{} addresses (1) by grouping Gaussians by spatial proximity and encoding structure-aware tokens that preserve global 3D structure for diffusion prior learning. To address (2), \MethodName{} factorizes the latent space into geometry and appearance, stabilizing parameter learning and enabling geometry-preserving edits, while still allowing appearance to leverage geometric context via controlled exchange rather than pure separation. Concretely, we use a dual-branch architecture with disentangled geometry/appearance latents and a lightweight cross-branch feature-sharing module that passes only necessary signals between branches.

\noindent\textbf{Encoding.} More formally, the input is a set of Gaussian splats $S = \left\{ S_{i}\right\}^{N}_{i=1}$, where after stacking the attributes we have $S = [\boldsymbol{\mu}, \mathbf{r}, \mathbf{s},  \mathbf{o}, \mathbf{c}] \in \mathbb{R}^{N \times 14}$, with each splat parameterized by its centroid  ${\boldsymbol{\mu}} \in \mathbb{R}^{N \times 3}$, scale $\mathbf{s} \in \mathbb{R}^{N \times 3}$, quaternion vector $\mathbf{r} \in \mathbb{R}^{N \times 4}$, opacity $\mathbf{o} \in \mathbb{R}^{N \times 1}$ and color $\mathbf{c} \in \mathbb{R}^{N \times 3}$, representing a 3D asset. First $S$ is randomly downsampled to a fixed number of splats, yielding $\hat{S} \in \mathbb{R}^{\hat{N} \times 14}$. We form local Gaussian groups via FPS+KNN, and encode them with a grouped-token transformer backbone as in \cite{shapesplat}. Specifically, we compute group centers $\boldsymbol{\mu_{\text{group}}} = \text{FPS}(\hat{\boldsymbol{\mu}}) \in \mathbb{R}^{p \times 3}$, where $p$ is the number of groups. For each $\boldsymbol{\mu_{\text{group}}}$, the $k$ nearest splats are obtained as $G = \text{KNN}(\hat{S}, \boldsymbol{\mu_{\text{group}}}) \in \mathbb{R}^{p \times k \times 14}$. The resulting Gaussian groups $G$ are then partitioned into $G_{g} \in \mathbb{R}^{p \times k \times 10}$ and $G_{c} \in \mathbb{R}^{p \times k \times 4}$, separating geometry (\emph{i.e.}, mean, quaternion vector, scale) and appearance (\emph{i.e.}, color, opacity) parameters. $G_{g}$ and $G_{c}$ are fed into tokenizers to obtain group tokens:

 \vspace{-1.3em}
\begin{equation}
T_g = \mathrm{Tokenizer}_g(G_g), \quad T_c = \mathrm{Tokenizer}_c(G_c)
\end{equation}

\noindent
Both \(T_g\) and \(T_c\) lie in \(\mathbb{R}^{1024\times 1024}\). Finally, the tokens are processed by the geometry and appearance encoders ${\mathcal{E}_{g}}$ and $ {\mathcal{E}_{c}}$ in our dual-branch VAE formulation:

 \vspace{-0.9em}
\begin{equation}
\label{encoderreparam}
    \mathbf{z}_{g}^{s} = {\mathcal{E}_{g}}(T_{g}), \quad  \mathbf{z}_{c}^{s} = {\mathcal{E}_{c}}(T_{c}), \quad \mathbf{z}_{g}^{s}, \mathbf{z}_{c}^{s} \in \mathbb{R}^{1024 \times 1024},
\end{equation}
 \noindent Here the encoders predict posterior parameters $(\boldsymbol{\mu}_g,\log\boldsymbol{\sigma}_g^2)$ and $(\boldsymbol{\mu}_c,\log\boldsymbol{\sigma}_c^2)$, from which we sample group-structured latents $\mathbf{z}_{g}^{s}$, $\mathbf{z}_{c}^{s}$ via the reparameterization trick; for brevity, we write this as \cref{encoderreparam}.
 
\noindent\textbf{Decoding.} To reconstruct parameters from $\mathbf{z}_{g}^{s}$ and $\mathbf{z}_{c}^{s}$, we employ dual-decoders with a lightweight cross-branch feature-sharing module $f_{\vartheta}$. The geometry decoder ${\mathcal{D}_{g}}$ predicts geometry tokens $T_{g}^{'}$ for centroids, scales and rotation quaternions, while the appearance decoder ${\mathcal{D}_{c}}$ produces appearance tokens $T_{c}^{'}$ for color and opacity. The transformer-based decoders are conditioned on positional embeddings $\varphi(\boldsymbol{\mu_{\text{group}}})$ derived from group centers to provide spatial context:
\begin{equation}
    T_{g}^{'} = {\mathcal{D}_{g}}(\mathbf{z}_{g}^{s}, \varphi(\boldsymbol{\mu_{\text{group}}})), \quad
    T_{c}^{'} = {\mathcal{D}_{c}}(\mathbf{z}_{c}^{s}, \varphi(\boldsymbol{\mu_{\text{group}}}))
\end{equation}

\noindent These tokens are then passed to the cross-branch feature-sharing module $f_{\vartheta}$, which uses bidirectional cross-attention to produce fused tokens $(T_{g}^{f}, T_{c}^{f}) = f_{\vartheta}(T_{g}^{'}, T_{c}^{'})$. The fused tokens are injected via a residual update:

 \vspace{-0.5em}
\begin{equation}
    \tilde{T}_{g} = T_{g}^{'} + \alpha T_{g}^{f}, \quad \tilde{T}_{c} = T_{c}^{'} + \alpha T_{c}^{f}
\end{equation}

\noindent Finally, a projection module $\Phi$ predicts the Gaussian attributes for the reconstructed groups $G'$:
\begin{equation}
    \boldsymbol{\mu'}, \mathbf{r'}, \mathbf{s'}, \mathbf{c'}, \mathbf{o'} = \Phi(\tilde{T}_{g}, \tilde{T}_{c})
\end{equation}

\noindent\textbf{Training.} To capture 3D priors, the Gaussian VAE is supervised by an $\mathcal{L}_{1}$ parameter reconstruction loss between input groups $G$ and reconstructions $G'$, denoted $\mathcal{L}_{param}$. However, because the mapping from image supervision to 3D Gaussian parameters is many-to-one, an $\mathcal{L}_{1}$ loss alone can be overly restrictive. Incorporating a render loss $\mathcal{L}_{render}$ relaxes this constraint and makes it possible to capture a broader range of valid solutions. The $\mathcal{L}_{1}$ render loss is computed over $V$ random views sampled from the training images used to optimize the input Gaussians and is complemented by an LPIPS loss \cite{perceptual}. 
 \vspace{-0.5em}
\begin{equation}
    \mathcal{L}_{render} = \mathcal{L}_{rgb}  + \tau\mathcal{L}_{LPIPS}
\end{equation}
Controlled sharing without entanglement is further enforced via a latent dissimilarity loss $\mathcal{L}_{LD}$. Latents are averaged over spatial positions and normalized, yielding ${\bar{\mathbf{z}}_g^s{(i)}}, {\bar{\mathbf{z}}_c^s{(i)}}$. Latent dissimilarity loss is computed as the mean cosine similarity across the batch $B$ to promote near-orthogonality:
 \vspace{-0.5em}
\begin{equation}
\mathcal{L}_{LD} = \frac{1}{B} \sum_{i=1}^{B} \left|\langle {\bar{\mathbf{z}}_g^s{(i)}}, {\bar{\mathbf{z}}_c^s{(i)}} \right\rangle|
\end{equation}

\noindent The Gaussian VAE is trained end-to-end with the following objective:
\begin{equation}
    \mathcal{L}_{GVAE} =  \lambda_{1}\mathcal{L}_{param} + \lambda_{2}\mathcal{L}_{render} + \lambda_{3}\mathcal{L}_{LD} + \lambda_{4}\mathcal{L}_{KL},
\end{equation}
\noindent where $\mathcal{L}_{\mathrm{KL}}
= D_{\mathrm{KL}}(q_g \,\|\, p) + D_{\mathrm{KL}}(q_c \,\|\, p)$ is the sum of KL divergence terms from the geometry and appearance posteriors to the Gaussian prior $p(z) = \mathcal{N}({0}, {I})$, scaled by a $\beta$-annealing schedule to stabilize latent-space learning.

\subsection{Stage 2: Latent Diffusion Pre-training}
\label{subsec:method2}


Our Gaussian VAE maps a 3DGS asset into a group-structured latent, we learn a text-conditioned diffusion prior in this latent space using a transformer denoiser. Concretely, we instantiate the denoiser with the architecture of Shap-E \cite{shapee} and initialize from its public weights, for stable, reproducible training, while remaining backbone-agnostic. Given an appearance latent $\mathbf{z}_{c}^{s}$, produced by ${\mathcal{E}_{c}}$ and a text embedding $\mathcal{C}$ (obtained by encoding text via CLIP \cite{clip}), the forward diffusion process gradually adds Gaussian noise to obtain a noisy latent ${\mathbf{z}_{c}^{s}}_\mathbf{t}$ at timestep $t\in\{1,\dots,T\}$.
The denoiser $d_{\varrho}$ is trained to recover the clean latent from ${\mathbf{z}_{c}^{s}}_\mathbf{t}$, conditioned on $\mathcal{C}$:

 \vspace{-1.0em}
\begin{equation}
\mathbb{E}_{{\mathbf{z}_{c}^{s}}, t, \epsilon \sim  \mathcal{N}(0, I)}\left\| d_{\varrho}({\mathbf{z}_{c}^{s}}_\mathbf{t}, \mathcal{C}, t) - {\mathbf{z}_{c}^{s}} \right\|^{2},    
\end{equation}

\noindent Here $\epsilon$ denotes the injected Gaussian noise sampled from $\mathcal{N}(0, I)$. Classifier-free guidance is also used. The diffusion loss is optimized jointly with the Stage-1 losses $\mathcal{L}_{param}$ and $\mathcal{L}_{render}$.

\subsection{Stage 3: 3D Appearance Stylization in Latent Space}
\label{subsec:method3}

Once the latent denoiser $d_{\varrho}$ is pre-trained, it is adapted to a latent editor that maps the source appearance latent $\mathbf{z}_c^s$ to an edited latent $\mathbf{z}_c^e = g(\mathbf{z}_c^s | \mathcal{C}^{e}, \mathbf{z}_g^s)$. Editing is supervised with a DMD-style distribution matching objective \cite{dmd1,dmd2} using a frozen 2D editor InstructPix2Pix (IP2P) \cite{ip2p} as teacher. For each training asset, $V$ rendered views of the current 3DGS are tiled into a multi-view grid, and encoded by the frozen SD VAE~\cite{stablediffusion} to obtain a latent $y$. A timestep $t$ is sampled and forward diffusion produces $y_t$. The \emph{teacher} (IP2P) predicts $\epsilon_{\mathrm{real}}(y_t,t)$ conditioned on edit text embedding $\mathcal{C}^{e}$, while an auxiliary denoiser $\epsilon_{\phi}$ is trained with the standard diffusion noise prediction loss on student samples under the same conditioning, yielding $\epsilon_{\mathrm{fake}}(y_t,t)$. Noise predictions are converted to clean-latent estimates $\hat{x}_0$, and the DMD update is applied in $\hat{x}_0$-space (with $\operatorname{sg}(\cdot)$ denoting the stop-gradient operator):

\begin{equation}\label{eq:dmd_stopgrad}\!
\begin{gathered}
\Delta=\frac{\hat{x}_{0,\mathrm{fake}}-\hat{x}_{0,\mathrm{real}}}{\gamma},
\qquad
\gamma=\operatorname{mean}_{c,h,w}\!\bigl|y-\hat{x}_{0,\mathrm{real}}\bigr|,\\[-0.2em]
\mathcal{L}_{\mathrm{DMD}}=\tfrac12\bigl\|y-\operatorname{sg}(y-\Delta)\bigr\|_2^2,
\qquad
\nabla_y\mathcal{L}_{\mathrm{DMD}}=\Delta.
\end{gathered}
\end{equation}

\noindent Since the DMD update can be unstable early when $\epsilon_{\phi}$ is inaccurate, we additionally apply a sparse $\mathcal{L}_1$ teacher-image anchoring loss, $\mathcal{L}_{MV\text{-}L1}$, between the IP2P-edited grid and the rendered views, and a color-variance regularizer $\mathcal{L}_{\mathrm{reg}}$ that enforces a minimum per-view RGB standard deviation (see supplementary for details and ablations). The final objective is:

\vspace{-0.3em}
\begin{equation}
\mathcal{L}_{\text{edit}}
=\mathcal{L}_{\mathrm{DMD}}
+\lambda_{MV\text{-}L1}\,\mathcal{L}_{MV\text{-}L1}
+\lambda_{\mathrm{reg}}\,\mathcal{L}_{\mathrm{reg}}
\end{equation}

\section{Experiments}
\label{experiments}

\begin{figure*}[!t]
\vspace{-1.0em}
  \centering
   \includegraphics[height=12.9cm]{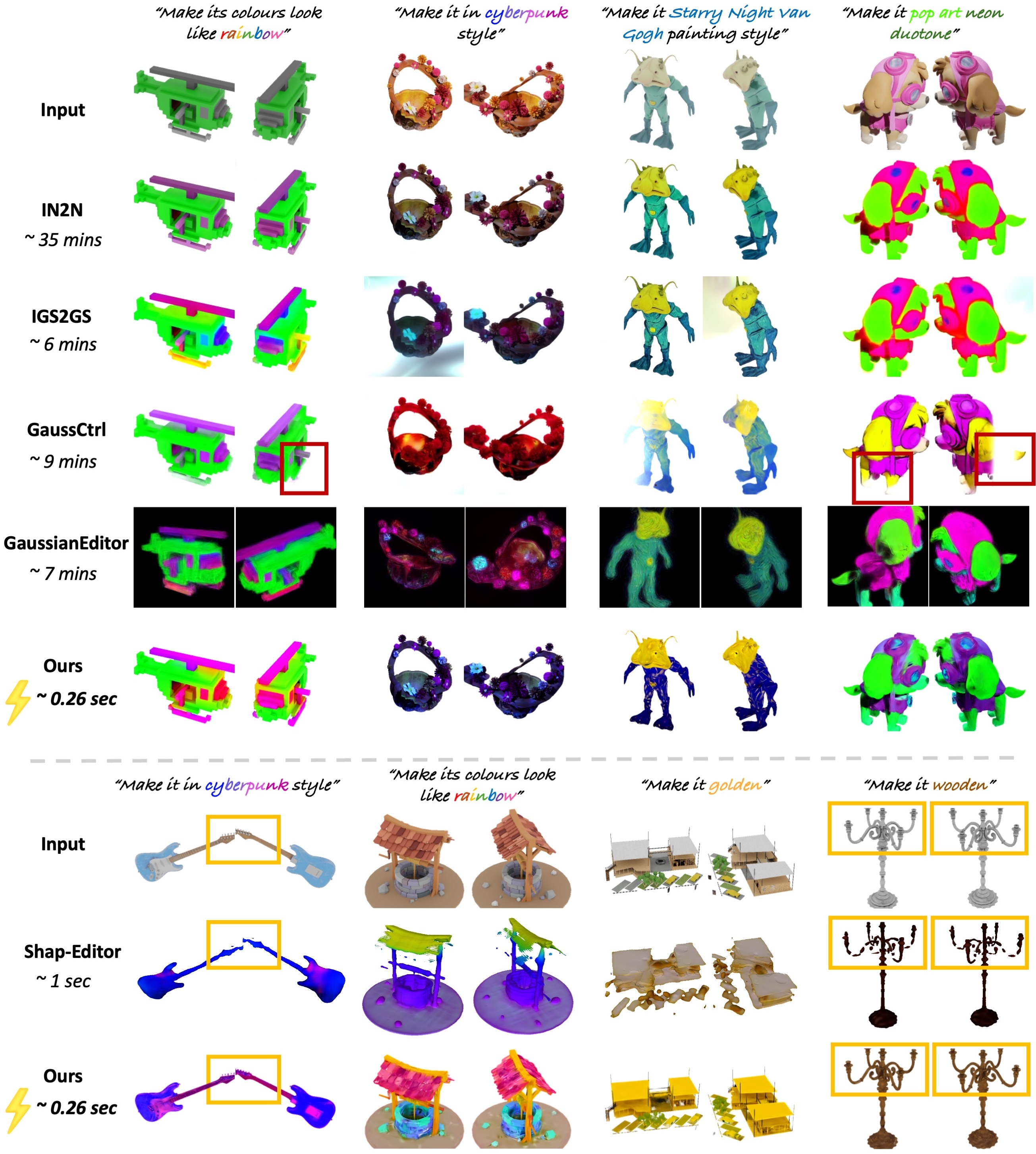}
   \caption{\textbf{Qualitative comparison with state-of-the-art methods.} 
Baselines often produce over-saturated, dramatic stylizations that distort 3D structure (\eg, blurred boundaries; rows~2,3,5, col.~4), minimal edits that are barely perceptible (row~2, col.~2,3), or severe geometric artifacts. In contrast, \MethodName{} delivers high-fidelity, text-aligned 3D stylizations with strong geometry preservation in a single feed-forward pass.}
  
   \label{fig:qual}
   \vspace{-1.5em}
\end{figure*}

\textbf{Dataset.} We use the TRELLIS-500K \cite{trellis} split of Objaverse \cite{objaverse}, filtered by aesthetic scores. From this split, we sample 140K 3D assets, render 72 views per asset, and generate 3DGS with LightGaussian~\cite{lightgaussian}. Assets with a PSNR below 30 dB are discarded from the dataset. The final filtered dataset contains approximately 123K 3D assets, of which 3K are randomly held out for testing purposes, and the remaining 120K are used for training. For latent diffusion pre-training, we use the split’s provided text prompts, for the editing stage, we curate a set of edit prompts.

\noindent \textbf{Evaluation Setup and Baselines.} Our approach is evaluated against five state-of-the-art text-conditioned 3D editing methods. IN2N \cite{in2n} represents assets with NeRFs, while IGS2GS \cite{igs2gs}, GaussianEditor \cite{gaussianeditor} and GaussCtrl \cite{gaussctrl} are 3DGS-based. All optimize the 3D representation by distilling from a 2D image editor. Shap-Editor \cite{shapeditor} is a feed-forward editing approach and, like ours, requires no test-time optimization. We evaluate on 30 test assets with 10 edit instructions against test-time optimization baselines, each instruction is applied to 3 assets. Since the training code for Shap-Editor is not publicly available, we evaluate it separately using its four released pre-trained models (one per edit instruction), applying each to 500 test assets (2000 total) sampled from our test split.

\noindent \textbf{Metrics.} Following prior work \cite{gaussctrl, ip2p, shapeditor, gaussianeditor}, we report the CLIP \cite{clip} similarity ($\mathrm{CLIP}_{\mathrm{sim}}$) and CLIP directional similarity ($\mathrm{CLIP}_{\mathrm{sim}}$) to evaluate alignment between the edit prompt and the resulting 3D edit. Following Shap-Editor \cite{shapeditor}, we also report Structure Distance (Structure Dist.) \cite{structuredistance} metric to quantify structural consistency between input and edited assets. We additionally conduct a user study to capture perceptual preferences.

\noindent\textbf{Implementation Details.} \MethodName{} is implemented in PyTorch \cite{pytorch}, and trained on 4 NVIDIA H100 GPUs with AdamW optimizer. The VAE uses LR $1\text{e}^{-3}$, weight decay $5\text{e}^{-2}$, CosLR scheduler with a 10-epoch warm-up, for 200 epochs. The diffusion model uses LR $1\text{e}{-5}$, the same weight decay and cosine annealing (200 epochs pretraining, 10 epochs editing). 

\subsection{Experimental Results}
\label{sec:experiments}

\noindent\textbf{Qualitative Comparison.}
Visual comparison of \MethodName{} with state-of-the-art 3D editing approaches is provided in \cref{fig:qual}. \MethodName{} produces high-fidelity 3D stylizations that surpass baseline methods in terms of text-alignment and geometry preservation in a single forward pass. GaussianEditor often yields noisy, sketch-like artifacts and alters geometry (\eg, blurred boundaries, structural loss; \textit{“Make it Starry Night Van Gogh style”}). IN2N, IGS2GS, and GaussCtrl tend to produce either over-saturated, dramatic edits that change also the 3D structure (\textit{“Make it pop-art neon duotone”}) or minimal edits that are barely perceptible (\textit{“Make its colors look like a rainbow”}). This is because optimization-based methods are sensitive to the 2D editor’s guidance scale and typically require careful per-asset tuning to avoid these extremes. In contrast, \MethodName{} is trained with guidance scales sampled from a reasonable range across many samples, yielding more realistic, geometry-preserving edits without tuning. Shap-Editor, by contrast, yields text-aligned 3D stylizations with a uniform edit strength across inputs, but often fails to preserve geometry, causing extreme smoothness, and structural distortion. This stems from editing a shared appearance–geometry representation, which limits control and entangles shape with style.

\noindent\textbf{Quantitative Comparison.} To evaluate text-alignment and structural consistency, we compute the average $\mathrm{CLIP}_{\mathrm{sim}}$, $\mathrm{CLIP}_{\mathrm{dir}}$ and Structure Dist. metric. As shown in \cref{tab:quantres}, \MethodName{} achieves the highest overall text-alignment and structure consistency. While IGS2GS \cite{igs2gs} and GaussianEditor \cite{gaussianeditor} report CLIP scores close to ours, these are inflated on scenes with dramatic, over-saturated color changes that exhibit severe multi-view inconsistency and geometric distortion. IN2N \cite{in2n} achieves a closer Structure Dist. by producing minimal edits, as reflected in its CLIP scores. Shap-Editor \cite{shapeditor} has the worst structural consistency as it introduces severe distortions.

\begin{table}[h]
\vspace{-1.0em}
  \centering
  \caption{\textbf{Quantitative comparison with state-of-the-art methods.} Our method achieves the best prompt–edit alignment and structural consistency while delivering stylized assets instantly.}
  \label{tab:quantres}
  \makebox[\textwidth][c]
    {
    \begin{tabularx}{\columnwidth}{lcccccc}
    \toprule
    \multicolumn{1}{l}{\textbf{Method}} & \multicolumn{1}{c}{\textbf{Optimization-free}}  & \multicolumn{3}{c}{\textbf{Evaluation Metric}} & \makecell[c]{\textbf{Inference}\\\textbf{Time $\downarrow$}}  \\

    \cmidrule{3-5}
    &  & \makecell{$\mathrm{CLIP}_{\mathrm{sim}}\!\uparrow$} & \makecell{$\mathrm{CLIP}_{\mathrm{dir}}\!\uparrow$} & \makecell{Structure \\ Dist. $\downarrow$} \\
    
    \hline
    IN2N \cite{in2n} & \ding{55} & 0.211 & 0.087 & 0.0085 & $\sim$ 35 mins  \\
    IGS2GS \cite{igs2gs}  & \ding{55} & 0.245 & 0.193 & 0.0356 & $\sim$ 6 mins \\
    GaussCtrl \cite{gaussctrl}  &\ding{55}& 0.231 & 0.159 & 0.0229 & $\sim$ 8 mins \\
    GausssianEditor \cite{gaussianeditor}  &\ding{55}& 0.246 & 0.189 & 0.0412 & $\sim$ 7 mins \\
    Shap-Editor \cite{shapeditor} & \checkmark & 0.238 & 0.128 & 0.0457 & $\sim$ 1 sec \\
    \hline
    Ours & \checkmark & \textbf{0.260} & \textbf{0.223} & \textbf{0.0050} & $\sim$ \textbf{0.26 sec}  \\
    \bottomrule
    \end{tabularx}
    }
    \vspace{-1.7em}
\end{table}

\begin{figure}[h]
\vspace{-1.8em}
  \centering
   \includegraphics[height=2.4cm]{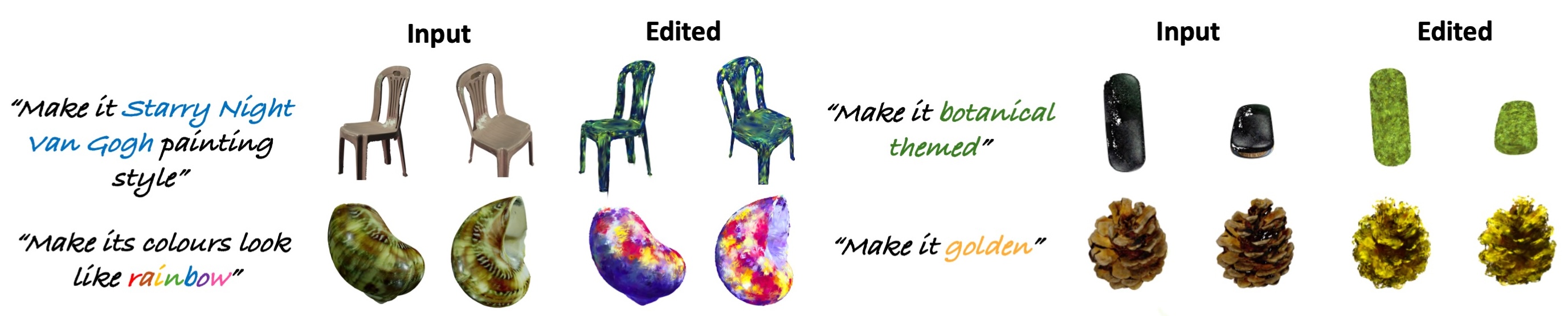}
   \caption{\textbf{Cross-dataset generalization on OmniObject3D \cite{omniobject3d}.} Our framework demonstrates strong style editing performance on out-of-distribution 3D assets.}
   \label{fig:generalization}
   \vspace{-1.7em}
\end{figure}

\noindent\textbf{Cross-data Generalization.} We evaluate on unseen objects from OmniObject3D~\cite{omniobject3d} (\cref{fig:generalization}) using 3DGS assets generated by LightGaussian~\cite{lightgaussian}. Despite OmniObject3D’s richer textures than Objaverse (our training data), \MethodName{} still generalizes out of distribution.

\begin{figure}[h]
\vspace{-1.0em}
  \centering
   \includegraphics[height=3.0cm]{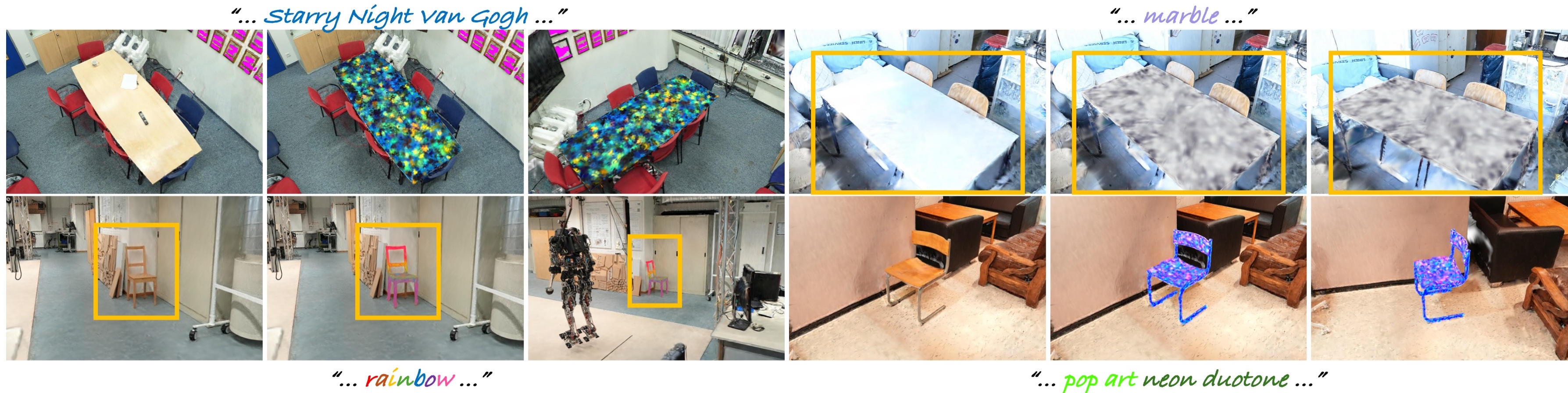}
   \caption{\textbf{Generalization to complex real-world scenes from SceneSplat \cite{scenesplat}.} Using a simple extract-edit-reinsert pipeline, \MethodName{} produces coherent stylizations despite capture artifacts, clutter, and occlusions.}
   \label{fig:scenesplat}
   \vspace{-1.9em}
\end{figure}

\label{text}
\noindent\textbf{Generalization to Complex Real-World Scenes.} In \cref{fig:scenesplat}, we show generalization to complex real-world scenes on the ScanNet++V2 split of SceneSplat~\cite{scenesplat} using the released 3DGS reconstructions. We localize the target object using per-Gaussian segmentation annotations, extract the corresponding subset of Gaussians, apply \MethodName{} to stylize the object, and then re-insert the edited Gaussians back into the original scene. This simple “extract–edit–reinsert” workflow makes \MethodName{} directly applicable to real-world 3DGS reconstructions, enabling object-level stylization in the presence of capture artifacts and clutter.

\noindent\textbf{User Study.} We conducted a user study with 50 participants to capture perceptual preferences. For each task, participants viewed outputs from \MethodName{} and the baselines alongside the edit prompt and the input 3D asset. Shap-Editor~\cite{shapeditor} was excluded because public checkpoints do not cover study's edit range. For each criterion, participants selected the output that (i) best aligns with the text prompt, (ii) best preserves the input asset’s 3D structure, and (iii) exhibits the highest visual quality. We also asked, conditioned on their preferred outputs, how long they would be willing to wait for a single asset. As summarized in \cref{tab:userstudy}, \MethodName{} was consistently preferred across all three criteria. In terms of latency, 69.6\% favored real-time ($<1$ s); the remaining choices were 2–10 mins (17.4\%), 10–20 mins (13.0\%), and 20–60 mins (0\%), indicating a clear preference for fast editing approaches.

\begin{table}
\vspace{-1.5em}
  \caption{\textbf{User study results.} \MethodName{} is consistently preferred over the baselines across the three criteria.}
  \label{tab:userstudy}
  \centering
  \begin{tabular}{@{}lccc@{}}
    \toprule
    Method & \makecell{Text\\Alignment $\uparrow$}
       & \makecell{Structural\\Consistency $\uparrow$}
       & \makecell{Visual\\Quality $\uparrow$} \\
    \midrule
    IN2N  &  16.30\% & 20.24\% & 21.68\% \\
    IGS2GS    & 14.67\% & 17.48\% & 20.21\% \\
    GaussCtrl  & 20.10\% & 15.84\% & 17.48\%\\
    GaussianEditor  & 15.21\% & 4.37\% & 6.01\%\\
    \hline
    Ours & \textbf{33.69\%} & \textbf{42.05\%} & \textbf{34.60\%}\\
    \bottomrule
  \end{tabular}
  \vspace{-1.7em}
\end{table}

\noindent\textbf{Inference Time.} Inference time denotes end-to-end latency from input receipt to final output. As shown in \cref{tab:quantres}, \MethodName{} produces edits in $\sim$ 0.26 second while achieving higher quality than baselines, enabling interactive use and large-scale production.

 \vspace{-1.2em}
\subsection{Ablation Study}

\noindent\textbf{Effect of Disentanglement on 3DGS Reconstruction.} We ablate the Gaussian VAE by evaluating standalone 3DGS parameter reconstruction performance. We compare a single-branch variant (\emph{w/o} disent.), which encodes appearance and geometry in a shared latent, against our dual-branch disentangled VAE (\emph{w/} disent. (Ours)) Both are trained only with the Stage 1 reconstruction objective. \cref{tab:quantevalgaussianvae} reports PSNR to ground-truth for LightGaussian, the input 3DGS, and reconstructed 3DGS renders, on 3K test samples. All assets use 16384 Gaussians.

\begin{figure}[h]
\vspace{-2.1em}
  \centering
   \includegraphics[height=2.1cm]{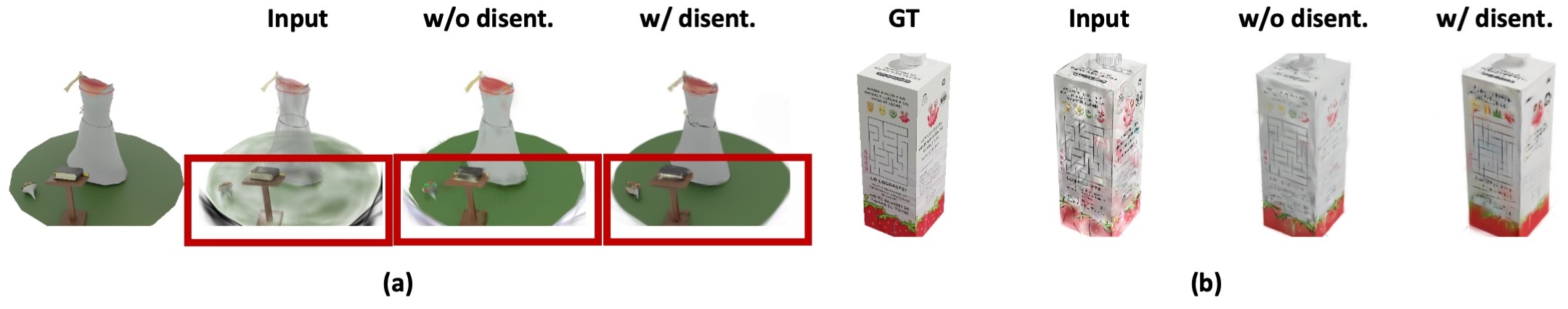}
   \caption{\textbf{Visual assessment of Gaussian VAE reconstructions.} The proposed dual-branch Gaussian VAE yields sharper reconstructions with better geometric fidelity and color accuracy.}
   \label{fig:reconst2}
   \vspace{-1.8em}
\end{figure}

Input 3DGS yields lower PSNR due to subsampling the original 50K Gaussians into a sparser representation, and inheriting LightGaussian \cite{lightgaussian} artifacts (\cref{fig:reconst2}a). With multi-view render supervision, both variants recover near ground-truth renders, mitigating sparsity and artifacts. Our GaussianVAE reaches high-fidelity reconstruction (PSNR $34.33$), only $\sim$0.88\,dB below the non-subsampled LightGaussian baseline (PSNR $35.21$), indicating minimal information loss. Disentanglement further improves quality, producing sharper reconstructions (\cref{fig:reconst2}b), higher geometric fidelity (\cref{fig:reconst2}a), and better appearance recovery; the non-disentangled variant tends to wash out colors (\cref{fig:reconst2}b).

\begin{table}[h]
\vspace{-1.2em}
\caption{\textbf{Effect of disentanglement on 3DGS parameter reconstruction.} \textbf{LightGaussian:} original LightGaussian~\cite{lightgaussian} reconstruction; \textbf{Input 3DGS}: downsampled LightGaussian 3DGS used as our model input; \textbf{\emph{w/o} disent.}: single-branch variant; \textbf{\emph{w/} disent.}: dual-branch Gaussian VAE.
}
\vspace{-0.5em}
\centering
\begin{tabular}{l|cccc}
\toprule
\multicolumn{1}{c|}{}  & \multicolumn{1}{c}{LightGaussian} & \multicolumn{1}{c}{Input 3DGS} & \multicolumn{1}{c}{\emph{w/o} disent.} & \multicolumn{1}{c}{\emph{w/} disent. (Ours)}  \\
\hline 
PSNR $\uparrow$ & 35.2117 & 32.7136  & 33.6635  &  34.3299  \\
\bottomrule
\end{tabular}
\label{tab:quantevalgaussianvae}
\vspace{-1.5em}
\end{table}

\noindent\textbf{Assessment of Framework Components.} We compare our full model against two variants (1) \emph{w/o} $\mathcal{L}_{LD}$, and (2) \emph{w/o} feat. sharing, which decodes Gaussian parameters directly from the two disentangled latents, omitting the cross-branch feature-sharing module. The quantitative results in \cref{tab:frameworkcomponents} confirm the effectiveness of $\mathcal{L}_{LD}$ and the feature-sharing module.

\vspace{-1.7em}
\begin{table}
  \caption{\textbf{Assessment of framework components.} We report $\mathrm{CLIP}_{\mathrm{sim}}\!\uparrow$ and $\mathrm{CLIP}_{\mathrm{dir}}\!\uparrow$ to compare different variants.}
  \label{tab:frameworkcomponents}
  \centering
  \begin{tabular}{@{}lcc@{}}
    \toprule
      & $\mathrm{CLIP}_{\mathrm{sim}}\!\uparrow$ & $\mathrm{CLIP}_{\mathrm{dir}}\!\uparrow$ \\
    \midrule
    (1) \emph{w/o} $\mathcal{L}_{LD}$  & 0.227 & 0.183 \\
    (2) \emph{w/o} feat. sharing  & 0.221 & 0.171 \\
    (3) Ours (full) & 0.260 & 0.223 \\
    \bottomrule
  \end{tabular}
  \vspace{-1.7em}
\end{table}

\begin{figure}[h]
  \centering
  \vspace{-1.0em}
   \includegraphics[width=0.7\linewidth]{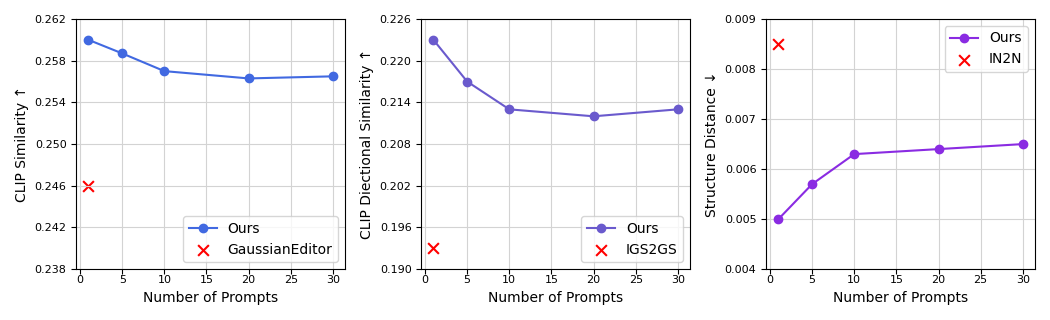}
   \caption{\textbf{Effect of style catalog size.} \MethodName{} maintains stylization performance as the style catalog size grows from 1 to 30.}
   \vspace{-2.0em}
   \label{fig:ablation_catlog}
\end{figure}

\noindent\textbf{Effect of Style Catalog Size.} In \cref{fig:ablation_catlog}, we vary the training prompt set (\ie catalog size) from 1 to 30 and evaluate on the same test set. Performance decreases slightly from 1 to 10 prompts as the training distribution broadens, and then stabilizes when scaling to 20–30 prompts; the 30-prompt model still outperforms the strongest baseline. This indicates \MethodName{} scales to larger style catalogs while maintaining stylization performance. See the supplementary for visuals.

\vspace{-1.5em}
\subsection{Application}

\begin{figure}[!h]
  \centering
   \vspace{-1.7em}
   \includegraphics[height=2.9cm]{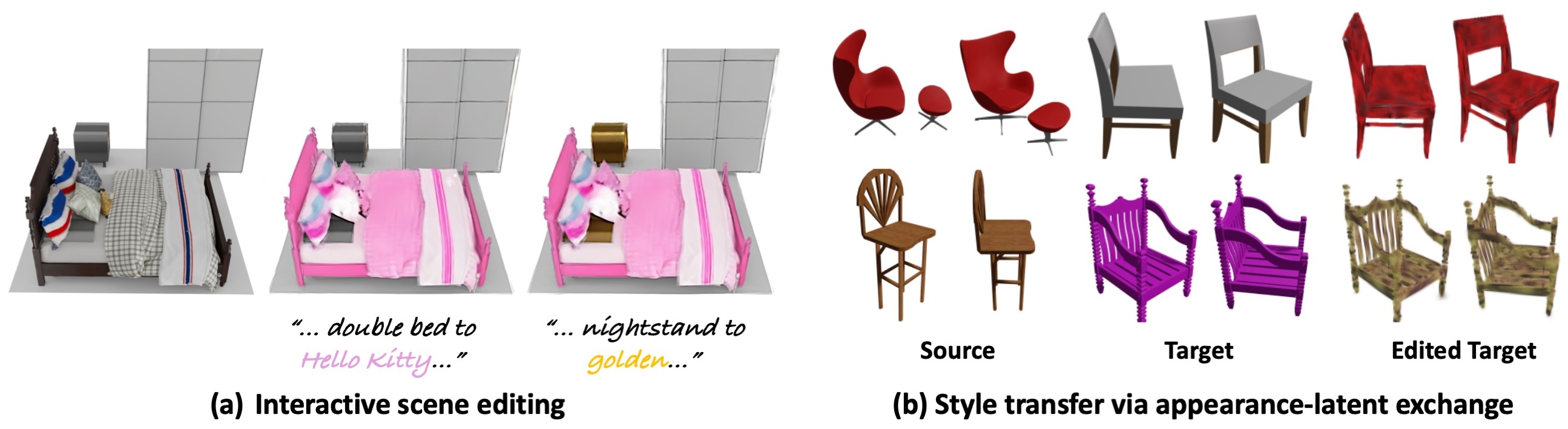}
   \caption{\textbf{Application.} \textbf{(a) Interactive scene editing:} \MethodName{} can be adopted for interactive object-level scene editing. \textbf{(b) Style transfer via appearance-latent exchange:}  Transferring the appearance code from a source to a target asset modifies appearance while preserving the geometric structure.}
   \label{fig:applications}
   \vspace{-1.7em}
\end{figure}

\noindent\textbf{Interactive Scene Editing.} Given a 3DGS scene and an edit prompt, we locate the target object using dataset-provided 3D bounding boxes (3D-FRONT \cite{3dfront1}), crop its 3DGS, and stylize it with \MethodName{}. Results (\cref{fig:teaser}, \cref{fig:applications}a) show our method can be adopted for interactive object-level scene editing.

\noindent\textbf{Style Transfer via Appearance-latent Exchange.} In \cref{fig:applications}b, appearance latents are transferred from source to target assets. This demonstrates controlled, geometry-preserving cross-asset appearance transfer enabled by our disentangled representation, while restricting cross-branch exchange to necessary signals. Despite direct latent swapping without post-processing, high-quality results are obtained.

\vspace{-0.7em}
\section{Conclusion}
\label{conclusion}
\vspace{-0.7em}

We introduce a diffusion-based, feed-forward 3DGS appearance stylizer that produces high-fidelity edits in a single pass, without test-time optimization. Operating in a structured, geometry-appearance disentangled latent space improves control, multi-view consistency, and geometry preservation. Experiments show \MethodName{} outperforms baselines in text-alignment and geometry preservation.

\noindent\textit{Limitations.} Similar to prior 2D-editor distillation methods, \MethodName{} inherits limitations of the its pre-trained components. We target curated, art-directed style catalogs and show in ablations that scaling to larger catalogs is feasible; achieving robust open-ended prompt coverage likely requires substantially larger training scale (\eg hundreds of GPUs) beyond our current compute budget.

\vspace{-1.0em}
\section*{Acknowledgements}
We thank members of the Bosch-UvA Delta Lab and anonymous reviewers for helpful discussions and feedback. This project was generously supported by the Bosch Center for Artificial Intelligence.

%
%
\bibliographystyle{splncs04}
\bibliography{main}

@String(CVPR  = {IEEE Conf. Comput. Vis. Pattern Recog.})

@String(ICCV  = {Int. Conf. Comput. Vis.})

@String(ECCV  = {Eur. Conf. Comput. Vis.})

@String(NeurIPS = {Adv. Neural Inform. Process. Syst.})

@String(ICML  = {Int. Conf. Mach. Learn.})

@String(ICLR  = {Int. Conf. Learn. Represent.})

@String(AAAI  = {AAAI})

@String(TOG   = {ACM Trans. Graph.})

@String(CVPR  = {CVPR})

@String(ICCV  = {ICCV})

@String(ECCV  = {ECCV})

@String(NeurIPS = {NeurIPS})

@String(ICML  = {ICML})

@String(ICLR  = {ICLR})

@String(TOG   = {ACM TOG})

@inproceedings{dreambooth,
  title={Dreambooth: Fine tuning text-to-image diffusion models for subject-driven generation},
  author={Ruiz, Nataniel and Li, Yuanzhen and Jampani, Varun and Pritch, Yael and Rubinstein, Michael and Aberman, Kfir},
  booktitle={Proceedings of the IEEE/CVF International Conference on Computer Vision (CVPR)},
  pages={22500--22510},
  year={2023}
}

@InProceedings{multiconceptcustom,
    author    = {Kumari, Nupur and Zhang, Bingliang and Zhang, Richard and Shechtman, Eli and Zhu, Jun-Yan},
    title     = {Multi-Concept Customization of Text-to-Image Diffusion},
    booktitle = {Proceedings of the IEEE/CVF Conference on Computer Vision and Pattern Recognition (CVPR)},
    month     = {June},
    year      = {2023},
    pages     = {1931-1941}
}

@article{animage,
  title={An image is worth one word: Personalizing text-to-image generation using textual inversion},
  author={Gal, Rinon and Alaluf, Yuval and Atzmon, Yuval and Patashnik, Or and Bermano, Amit H and Chechik, Gal and Cohen-Or, Daniel},
  journal={arXiv preprint arXiv:2208.01618},
  year={2022}
}

@InProceedings{dragfusion,
    author    = {Shi, Yujun and Xue, Chuhui and Liew, Jun Hao and Pan, Jiachun and Yan, Hanshu and Zhang, Wenqing and Tan, Vincent Y. F. and Bai, Song},
    title     = {DragDiffusion: Harnessing Diffusion Models for Interactive Point-based Image Editing},
    booktitle = {Proceedings of the IEEE/CVF Conference on Computer Vision and Pattern Recognition (CVPR)},
    month     = {June},
    year      = {2024},
    pages     = {8839-8849}
}

@article{mdragondiffusion,
  title={Dragondiffusion: Enabling drag-style manipulation on diffusion models},
  author={Mou, Chong and Wang, Xintao and Song, Jiechong and Shan, Ying and Zhang, Jian},
  journal={arXiv preprint arXiv:2307.02421},
  year={2023}
}

@article{layout1,
  title={Diffusion self-guidance for controllable image generation},
  author={Epstein, Dave and Jabri, Allan and Poole, Ben and Efros, Alexei and Holynski, Aleksander},
  journal={Advances in Neural Information Processing Systems (NeurIPS)},
  volume={36},
  pages={16222--16239},
  year={2023}
}

@InProceedings{layout2,
    author    = {Li, Yuheng and Liu, Haotian and Wu, Qingyang and Mu, Fangzhou and Yang, Jianwei and Gao, Jianfeng and Li, Chunyuan and Lee, Yong Jae},
    title     = {GLIGEN: Open-Set Grounded Text-to-Image Generation},
    booktitle = {Proceedings of the IEEE/CVF Conference on Computer Vision and Pattern Recognition (CVPR)},
    month     = {June},
    year      = {2023},
    pages     = {22511-22521}
}

@article{layout3,
  title={Attend-and-excite: Attention-based semantic guidance for text-to-image diffusion models},
  author={Chefer, Hila and Alaluf, Yuval and Vinker, Yael and Wolf, Lior and Cohen-Or, Daniel},
  journal={ACM transactions on Graphics (TOG)},
  volume={42},
  number={4},
  pages={1--10},
  year={2023},
  publisher={ACM New York, NY, USA}
}

@article{glide,
  title={Glide: Towards photorealistic image generation and editing with text-guided diffusion models},
  author={Nichol, Alex and Dhariwal, Prafulla and Ramesh, Aditya and Shyam, Pranav and Mishkin, Pamela and McGrew, Bob and Sutskever, Ilya and Chen, Mark},
  journal={arXiv preprint arXiv:2112.10741},
  year={2021}
}

@inproceedings{controlnet,
  title={Adding conditional control to text-to-image diffusion models},
  author={Zhang, Lvmin and Rao, Anyi and Agrawala, Maneesh},
  booktitle={Proceedings of the IEEE/CVF international Conference on Computer Vision (ICCV)},
  pages={3836--3847},
  year={2023}
}

@inproceedings{text2live,
  title={Text2live: Text-driven layered image and video editing},
  author={Bar-Tal, Omer and Ofri-Amar, Dolev and Fridman, Rafail and Kasten, Yoni and Dekel, Tali},
  booktitle={European Conference on Computer Vision (ECCV)},
  pages={707--723},
  year={2022},
  organization={Springer}
}

@article{p2p,
  title={Prompt-to-prompt image editing with cross attention control},
  author={Hertz, Amir and Mokady, Ron and Tenenbaum, Jay and Aberman, Kfir and Pritch, Yael and Cohen-Or, Daniel},
  journal={arXiv preprint arXiv:2208.01626},
  year={2022}
}

@inproceedings{imagic,
  title={Imagic: Text-based real image editing with diffusion models},
  author={Kawar, Bahjat and Zada, Shiran and Lang, Oran and Tov, Omer and Chang, Huiwen and Dekel, Tali and Mosseri, Inbar and Irani, Michal},
  booktitle={Proceedings of the IEEE/CVF Conference on Computer Vision and Pattern Recognition (CVPR)},
  pages={6007--6017},
  year={2023}
}

@article{sdedit,
  title={Sdedit: Guided image synthesis and editing with stochastic differential equations},
  author={Meng, Chenlin and He, Yutong and Song, Yang and Song, Jiaming and Wu, Jiajun and Zhu, Jun-Yan and Ermon, Stefano},
  journal={ICLR},
  year={2022}
}

@InProceedings{nulltext,
    author    = {Mokady, Ron and Hertz, Amir and Aberman, Kfir and Pritch, Yael and Cohen-Or, Daniel},
    title     = {NULL-Text Inversion for Editing Real Images Using Guided Diffusion Models},
    booktitle = {Proceedings of the IEEE/CVF Conference on Computer Vision and Pattern Recognition (CVPR)},
    month     = {June},
    year      = {2023},
    pages     = {6038-6047}
}

@inproceedings{zeroshoti2i,
  title={Zero-shot image-to-image translation},
  author={Parmar, Gaurav and Kumar Singh, Krishna and Zhang, Richard and Li, Yijun and Lu, Jingwan and Zhu, Jun-Yan},
  booktitle={ACM SIGGRAPH 2023 conference proceedings},
  pages={1--11},
  year={2023}
}

@InProceedings{plugandplay,
    author    = {Tumanyan, Narek and Geyer, Michal and Bagon, Shai and Dekel, Tali},
    title     = {Plug-and-Play Diffusion Features for Text-Driven Image-to-Image Translation},
    booktitle = {Proceedings of the IEEE/CVF Conference on Computer Vision and Pattern Recognition (CVPR)},
    month     = {June},
    year      = {2023},
    pages     = {1921-1930}
}

@InProceedings{ip2p,
    author    = {Brooks, Tim and Holynski, Aleksander and Efros, Alexei A.},
    title     = {InstructPix2Pix: Learning To Follow Image Editing Instructions},
    booktitle = {Proceedings of the IEEE/CVF Conference on Computer Vision and Pattern Recognition (CVPR)},
    month     = {June},
    year      = {2023},
    pages     = {18392-18402}
}

@InProceedings{hive,
    author    = {Zhang, Shu and Yang, Xinyi and Feng, Yihao and Qin, Can and Chen, Chia-Chih and Yu, Ning and Chen, Zeyuan and Wang, Huan and Savarese, Silvio and Ermon, Stefano and Xiong, Caiming and Xu, Ran},
    title     = {HIVE: Harnessing Human Feedback for Instructional Visual Editing},
    booktitle = {Proceedings of the IEEE/CVF Conference on Computer Vision and Pattern Recognition (CVPR)},
    month     = {June},
    year      = {2024},
    pages     = {9026-9036}
}

@inproceedings{multireward,
  author={Xin Gu and Ming Li and Libo Zhang and Fan Chen and Longyin Wen and Tiejian Luo and Sijie Zhu},
  title={Multi-Reward as Condition for Instruction-based Image Editing},
  booktitle={International Conference on Learning Representations (ICLR)},
  year={2025}
}

@inproceedings{watchyoursteps,
  title={Watch your steps: Local image and scene editing by text instructions},
  author={Mirzaei, Ashkan and Aumentado-Armstrong, Tristan and Brubaker, Marcus A and Kelly, Jonathan and Levinshtein, Alex and Derpanis, Konstantinos G and Gilitschenski, Igor},
  booktitle={European Conference on Computer Vision (ECCV)},
  pages={111--129},
  year={2024},
  organization={Springer}
}

@article{tango,
  title={Tango: Text-driven photorealistic and robust 3d stylization via lighting decomposition},
  author={Chen, Yongwei and Chen, Rui and Lei, Jiabao and Zhang, Yabin and Jia, Kui},
  journal={Advances in Neural Information Processing Systems (NeurIPS)},
  volume={35},
  pages={30923--30936},
  year={2022}
}

@InProceedings{text2mesh,
    author    = {Michel, Oscar and Bar-On, Roi and Liu, Richard and Benaim, Sagie and Hanocka, Rana},
    title     = {Text2Mesh: Text-Driven Neural Stylization for Meshes},
    booktitle = {Proceedings of the IEEE/CVF Conference on Computer Vision and Pattern Recognition (CVPR)},
    month     = {June},
    year      = {2022},
    pages     = {13492-13502}
}

@InProceedings{clipnerf,
    author    = {Wang, Can and Chai, Menglei and He, Mingming and Chen, Dongdong and Liao, Jing},
    title     = {CLIP-NeRF: Text-and-Image Driven Manipulation of Neural Radiance Fields},
    booktitle = {Proceedings of the IEEE/CVF Conference on Computer Vision and Pattern Recognition (CVPR)},
    month     = {June},
    year      = {2022},
    pages     = {3835-3844}
}

@article{nerfart,
  title={Nerf-art: Text-driven neural radiance fields stylization},
  author={Wang, Can and Jiang, Ruixiang and Chai, Menglei and He, Mingming and Chen, Dongdong and Liao, Jing},
  journal={IEEE Transactions on Visualization and Computer Graphics},
  year={2023},
  publisher={IEEE}
}

@InProceedings{temo,
    author    = {Zhang, Xuying and Yin, Bo-Wen and Chen, Yuming and Lin, Zheng and Li, Yunheng and Hou, Qibin and Cheng, Ming-Ming},
    title     = {TeMO: Towards Text-Driven 3D Stylization for Multi-Object Meshes},
    booktitle = {Proceedings of the IEEE/CVF Conference on Computer Vision and Pattern Recognition (CVPR)},
    month     = {June},
    year      = {2024},
    pages     = {19531-19540}
}

@InProceedings{in2n,
    author    = {Haque, Ayaan and Tancik, Matthew and Efros, Alexei A. and Holynski, Aleksander and Kanazawa, Angjoo},
    title     = {Instruct-NeRF2NeRF: Editing 3D Scenes with Instructions},
    booktitle = {Proceedings of the IEEE/CVF International Conference on Computer Vision (ICCV)},
    month     = {October},
    year      = {2023},
    pages     = {19740-19750}
}

@article{vicanerf,
  title={Vica-nerf: View-consistency-aware 3d editing of neural radiance fields},
  author={Dong, Jiahua and Wang, Yu-Xiong},
  journal={Advances in Neural Information Processing Systems (NeurIPS)},
  volume={36},
  pages={61466--61477},
  year={2023}
}

@article{Instructp2p,
  title={Instructp2p: Learning to edit 3d point clouds with text instructions},
  author={Xu, Jiale and Wang, Xintao and Cao, Yan-Pei and Cheng, Weihao and Shan, Ying and Gao, Shenghua},
  journal={arXiv preprint arXiv:2306.07154},
  year={2023}
}

@misc{igs2gs,
  title={Instruct-GS2GS: Editing 3D Gaussian Splats with Instructions},
  author={Vachha, Cyrus and Haque, Ayaan},
  year={2024}
}

@InProceedings{gaussianeditor,
    author    = {Chen, Yiwen and Chen, Zilong and Zhang, Chi and Wang, Feng and Yang, Xiaofeng and Wang, Yikai and Cai, Zhongang and Yang, Lei and Liu, Huaping and Lin, Guosheng},
    title     = {GaussianEditor: Swift and Controllable 3D Editing with Gaussian Splatting},
    booktitle = {Proceedings of the IEEE/CVF Conference on Computer Vision and Pattern Recognition (CVPR)},
    month     = {June},
    year      = {2024},
    pages     = {21476-21485}
}

@article{proteusnerf,
  title={Proteusnerf: Fast lightweight nerf editing using 3d-aware image context},
  author={Wang, Binglun and Dutt, Niladri Shekhar and Mitra, Niloy J},
  journal={Proceedings of the ACM on Computer Graphics and Interactive Techniques},
  volume={7},
  number={1},
  pages={1--17},
  year={2024},
  publisher={ACM New York, NY, USA}
}

@inproceedings{editsplat,
  title={Editsplat: Multi-view fusion and attention-guided optimization for view-consistent 3d scene editing with 3d gaussian splatting},
  author={Lee, Dong In and Park, Hyeongcheol and Seo, Jiyoung and Park, Eunbyung and Park, Hyunje and Baek, Ha Dam and Shin, Sangheon and Kim, Sangmin and Kim, Sangpil},
  booktitle={Proceedings of the Computer Vision and Pattern Recognition Conference (CVPR)},
  pages={11135--11145},
  year={2025}
}

@inproceedings{gaussctrl,
  title={Gaussctrl: Multi-view consistent text-driven 3d gaussian splatting editing},
  author={Wu, Jing and Bian, Jia-Wang and Li, Xinghui and Wang, Guangrun and Reid, Ian and Torr, Philip and Prisacariu, Victor Adrian},
  booktitle={European Conference on Computer Vision (ECCV)},
  pages={55--71},
  year={2024},
  organization={Springer}
}

@inproceedings{dge,
  title={Dge: Direct gaussian 3d editing by consistent multi-view editing},
  author={Chen, Minghao and Laina, Iro and Vedaldi, Andrea},
  booktitle={European Conference on Computer Vision (ECCV)},
  pages={74--92},
  year={2024},
  organization={Springer}
}

@InProceedings{consistentdreamer,
    author    = {Chen, Jun-Kun and Bul\`o, Samuel Rota and M\"uller, Norman and Porzi, Lorenzo and Kontschieder, Peter and Wang, Yu-Xiong},
    title     = {ConsistDreamer: 3D-Consistent 2D Diffusion for High-Fidelity Scene Editing},
    booktitle = {Proceedings of the IEEE/CVF Conference on Computer Vision and Pattern Recognition (CVPR)},
    month     = {June},
    year      = {2024},
    pages     = {21071-21080}
}

@InProceedings{morpheus,
  title={Morpheus: Text-Driven 3D Gaussian Splat Shape and Color Stylization},
  author={Wynn, Jamie and Qureshi, Zawar and Powierza, Jakub and Watson, Jamie and Sayed, Mohamed},
  booktitle={Proceedings of the IEEE/CVF Conference on Computer Vision and Pattern Recognition (CVPR)},
  year={2025}
}

@inproceedings{viewconsistent3d,
  title={View-consistent 3d editing with gaussian splatting},
  author={Wang, Yuxuan and Yi, Xuanyu and Wu, Zike and Zhao, Na and Chen, Long and Zhang, Hanwang},
  booktitle={European Conference on Computer Vision (ECCV)},
  pages={404--420},
  year={2024},
  organization={Springer}
}

@article{instruct3d23d,
  title={Instruct 3d-to-3d: Text instruction guided 3d-to-3d conversion},
  author={Kamata, Hiromichi and Sakuma, Yuiko and Hayakawa, Akio and Ishii, Masato and Narihira, Takuya},
  journal={arXiv preprint arXiv:2303.15780},
  year={2023}
}

@inproceedings{ednerf,
  author={JangHo Park and Gihyun Kwon and Jong Chul Ye},
  title={{ED-NeRF:} Efficient Text-Guided Editing of 3D Scene With Latent Space {NeRF}},
  booktitle={International Conference on Learning Representations (ICLR)},
  year={2024}
}

@InProceedings{voxe,
    author    = {Sella, Etai and Fiebelman, Gal and Hedman, Peter and Averbuch-Elor, Hadar},
    title     = {Vox-E: Text-Guided Voxel Editing of 3D Objects},
    booktitle = {Proceedings of the IEEE/CVF International Conference on Computer Vision (ICCV)},
    month     = {October},
    year      = {2023},
    pages     = {430-440}
}

@inproceedings{dreameditor,
  title={Dreameditor: Text-driven 3d scene editing with neural fields},
  author={Zhuang, Jingyu and Wang, Chen and Lin, Liang and Liu, Lingjie and Li, Guanbin},
  booktitle={SIGGRAPH Asia 2023 Conference Papers},
  pages={1--10},
  year={2023}
}

@inproceedings{focaldreamer,
  title={Focaldreamer: Text-driven 3d editing via focal-fusion assembly},
  author={Li, Yuhan and Dou, Yishun and Shi, Yue and Lei, Yu and Chen, Xuanhong and Zhang, Yi and Zhou, Peng and Ni, Bingbing},
  booktitle={Proceedings of the AAAI conference on artificial intelligence},
  volume={38},
  number={4},
  pages={3279--3287},
  year={2024}
}

@InProceedings{customizenerf,
    author    = {He, Runze and Huang, Shaofei and Nie, Xuecheng and Hui, Tianrui and Liu, Luoqi and Dai, Jiao and Han, Jizhong and Li, Guanbin and Liu, Si},
    title     = {Customize your NeRF: Adaptive Source Driven 3D Scene Editing via Local-Global Iterative Training},
    booktitle = {Proceedings of the IEEE/CVF Conference on Computer Vision and Pattern Recognition (CVPR)},
    month     = {June},
    year      = {2024},
    pages     = {6966-6975}
}

@article{editdiffnerf,
  title={Edit-diffnerf: Editing 3d neural radiance fields using 2d diffusion model},
  author={Yu, Lu and Xiang, Wei and Han, Kang},
  journal={arXiv preprint arXiv:2306.09551},
  year={2023}
}

@inproceedings{freeditor,
  title={Free-editor: zero-shot text-driven 3D scene editing},
  author={Karim, Nazmul and Iqbal, Hasan and Khalid, Umar and Chen, Chen and Hua, Jing},
  booktitle={European Conference on Computer Visio (ECCV)},
  pages={436--453},
  year={2024},
  organization={Springer}
}

@InProceedings{shapeditor,
    author    = {Chen, Minghao and Xie, Junyu and Laina, Iro and Vedaldi, Andrea},
    title     = {SHAP-EDITOR: Instruction-Guided Latent 3D Editing in Seconds},
    booktitle = {Proceedings of the IEEE/CVF Conference on Computer Vision and Pattern Recognition (CVPR)},
    month     = {June},
    year      = {2024},
    pages     = {26456-26466}
}

@article{pytorch,
  title={High-Performance Deep Learning Library},
  author={Style, PyTorch An Imperative},
  journal={Advances in Neural},
  year={2024}
}

@inproceedings{magic3d,
  title={Magic3d: High-resolution text-to-3d content creation},
  author={Lin, Chen-Hsuan and Gao, Jun and Tang, Luming and Takikawa, Towaki and Zeng, Xiaohui and Huang, Xun and Kreis, Karsten and Fidler, Sanja and Liu, Ming-Yu and Lin, Tsung-Yi},
  booktitle={Proceedings of the IEEE/CVF Conference on Computer Vision and Pattern Recognition (CVPR)},
  pages={300--309},
  year={2023}
}

@Article{gaussiansplatting,
      author       = {Kerbl, Bernhard and Kopanas, Georgios and Leimk{\"u}hler, Thomas and Drettakis, George},
      title        = {3D Gaussian Splatting for Real-Time Radiance Field Rendering},
      journal      = {ACM Transactions on Graphics},
      number       = {4},
      volume       = {42},
      month        = {July},
      year         = {2023},
      url          = {https://repo-sam.inria.fr/fungraph/3d-gaussian-splatting/}
}

@article{ddpm,
  title={Denoising diffusion probabilistic models},
  author={Ho, Jonathan and Jain, Ajay and Abbeel, Pieter},
  journal={Advances in neural information processing systems (NeurIPS)},
  volume={33},
  pages={6840--6851},
  year={2020}
}

@inproceedings{stablediffusion,
  title={High-resolution image synthesis with latent diffusion models},
  author={Rombach, Robin and Blattmann, Andreas and Lorenz, Dominik and Esser, Patrick and Ommer, Bj{\"o}rn},
  booktitle={Proceedings of the IEEE/CVF Conference on Computer Vision and Pattern Recognition (CVPR)},
  pages={10684--10695},
  year={2022}
}

@inproceedings{dreamfusion,
title={DreamFusion: Text-to-3D using 2D Diffusion},
author={Ben Poole and Ajay Jain and Jonathan T. Barron and Ben Mildenhall},
booktitle={The Eleventh International Conference on Learning Representations (ICLR)},
year={2023},
url={https://openreview.net/forum?id=FjNys5c7VyY}
}

@InProceedings{fantasia3d,
author    = {Chen, Rui and Chen, Yongwei and Jiao, Ningxin and Jia, Kui},
title     = {Fantasia3D: Disentangling Geometry and Appearance for High-quality Text-to-3D Content Creation},
booktitle = {Proceedings of the IEEE/CVF International Conference on Computer Vision (ICCV)},
month     = {October},
year      = {2023}
}

@article{sweetdreamer,
  title={Sweetdreamer: Aligning geometric priors in 2d diffusion for consistent text-to-3d},
  author={Li, Weiyu and Chen, Rui and Chen, Xuelin and Tan, Ping},
  journal={arXiv preprint arXiv:2310.02596},
  year={2023}
}

@inproceedings{gaussiandreamer,
  title={Gaussiandreamer: Fast generation from text to 3d gaussians by bridging 2d and 3d diffusion models},
  author={Yi, Taoran and Fang, Jiemin and Wang, Junjie and Wu, Guanjun and Xie, Lingxi and Zhang, Xiaopeng and Liu, Wenyu and Tian, Qi and Wang, Xinggang},
  booktitle={Proceedings of the IEEE/CVF Conference on Computer Vision and Pattern Recognition (CVPR)},
  pages={6796--6807},
  year={2024}
}

@article{dreamgaussian,
  title={Dreamgaussian: Generative gaussian splatting for efficient 3d content creation},
  author={Tang, Jiaxiang and Ren, Jiawei and Zhou, Hang and Liu, Ziwei and Zeng, Gang},
  journal={arXiv preprint arXiv:2309.16653},
  year={2023}
}

@inproceedings{zero123,
  title={Zero-1-to-3: Zero-shot one image to 3d object},
  author={Liu, Ruoshi and Wu, Rundi and Van Hoorick, Basile and Tokmakov, Pavel and Zakharov, Sergey and Vondrick, Carl},
  booktitle={Proceedings of the IEEE/CVF Conference on Computer Vision and Pattern Recognition (CVPR)},
  pages={9298--9309},
  year={2023}
}

@article{mvdream,
  title={Mvdream: Multi-view diffusion for 3d generation},
  author={Shi, Yichun and Wang, Peng and Ye, Jianglong and Long, Mai and Li, Kejie and Yang, Xiao},
  journal={arXiv preprint arXiv:2308.16512},
  year={2023}
}

@article{mvdiffusion,
  author = {Tang, Shitao and Zhang, Fuyang and Chen, Jiacheng and Wang, Peng and Furukawa, Yasutaka},
  title = {MVDiffusion: Enabling Holistic Multi-view Image Generation with Correspondence-Aware Diffusion},
  journal = {arXiv},
  year = {2023},
}

@article{syncdreamer,
  title={Syncdreamer: Generating multiview-consistent images from a single-view image},
  author={Liu, Yuan and Lin, Cheng and Zeng, Zijiao and Long, Xiaoxiao and Liu, Lingjie and Komura, Taku and Wang, Wenping},
  journal={arXiv preprint arXiv:2309.03453},
  year={2023}
}

@inproceedings{wonder3d,
  title={Wonder3d: Single image to 3d using cross-domain diffusion},
  author={Long, Xiaoxiao and Guo, Yuan-Chen and Lin, Cheng and Liu, Yuan and Dou, Zhiyang and Liu, Lingjie and Ma, Yuexin and Zhang, Song-Hai and Habermann, Marc and Theobalt, Christian and others},
  booktitle={Proceedings of the IEEE/CVF Conference on Computer Vision and Pattern Recognition (CVPR)},
  pages={9970--9980},
  year={2024}
}

@article{instant3d,
  title={Instant3d: Fast text-to-3d with sparse-view generation and large reconstruction model},
  author={Li, Jiahao and Tan, Hao and Zhang, Kai and Xu, Zexiang and Luan, Fujun and Xu, Yinghao and Hong, Yicong and Sunkavalli, Kalyan and Shakhnarovich, Greg and Bi, Sai},
  journal={arXiv preprint arXiv:2311.06214},
  year={2023}
}

@inproceedings{lgm,
  title={Lgm: Large multi-view gaussian model for high-resolution 3d content creation},
  author={Tang, Jiaxiang and Chen, Zhaoxi and Chen, Xiaokang and Wang, Tengfei and Zeng, Gang and Liu, Ziwei},
  booktitle={European Conference on Computer Vision (ECCV)},
  pages={1--18},
  year={2024},
  organization={Springer}
}

@inproceedings{diffsplat,
  title={DiffSplat: Repurposing Image Diffusion Models for Scalable 3D Gaussian Splat Generation},
  author={Lin, Chenguo and Pan, Panwang and Yang, Bangbang and Li, Zeming and Mu, Yadong},
  booktitle={International Conference on Learning Representations (ICLR)},
  year={2025}
}

@inproceedings{gvgen,
  title={Gvgen: Text-to-3d generation with volumetric representation},
  author={He, Xianglong and Chen, Junyi and Peng, Sida and Huang, Di and Li, Yangguang and Huang, Xiaoshui and Yuan, Chun and Ouyang, Wanli and He, Tong},
  booktitle={European Conference on Computer Vision},
  pages={463--479},
  year={2024},
  organization={Springer}
}

@inproceedings{trellis,
  title={Structured 3d latents for scalable and versatile 3d generation},
  author={Xiang, Jianfeng and Lv, Zelong and Xu, Sicheng and Deng, Yu and Wang, Ruicheng and Zhang, Bowen and Chen, Dong and Tong, Xin and Yang, Jiaolong},
  booktitle={Proceedings of the Computer Vision and Pattern Recognition Conference (CVPR)},
  pages={21469--21480},
  year={2025}
}

@article{clay,
  title={CLAY: A Controllable Large-scale Generative Model for Creating High-quality 3D Assets},
  author={Zhang, Longwen and Wang, Ziyu and Zhang, Qixuan and Qiu, Qiwei and Pang, Anqi and Jiang, Haoran and Yang, Wei and Xu, Lan and Yu, Jingyi},
  journal={ACM Transactions on Graphics (TOG)},
  volume={43},
  number={4},
  pages={1--20},
  year={2024},
  publisher={ACM New York, NY, USA}
}

@InProceedings{3dtopia,
    author    = {Chen, Zhaoxi and Tang, Jiaxiang and Dong, Yuhao and Cao, Ziang and Hong, Fangzhou and Lan, Yushi and Wang, Tengfei and Xie, Haozhe and Wu, Tong and Saito, Shunsuke and Pan, Liang and Lin, Dahua and Liu, Ziwei},
    title     = {3DTopia-XL: Scaling High-quality 3D Asset Generation via Primitive Diffusion},
    booktitle = {Proceedings of the IEEE/CVF Conference on Computer Vision and Pattern Recognition (CVPR)},
    month     = {June},
    year      = {2025},
    pages     = {26576-26586}
}

@article{atlas,
  title={Atlas Gaussians Diffusion for 3D Generation},
  author={Yang, Haitao and Dong, Yuan and Jiang, Hanwen and Xu, Dejia and Pavlakos, Georgios and Huang, Qixing},
  journal={arXiv preprint arXiv:2408.13055},
  year={2024}
}

@inproceedings{gpld3d,
  title={Gpld3d: Latent diffusion of 3d shape generative models by enforcing geometric and physical priors},
  author={Dong, Yuan and Zuo, Qi and Gu, Xiaodong and Yuan, Weihao and Zhao, Zhengyi and Dong, Zilong and Bo, Liefeng and Huang, Qixing},
  booktitle={Proceedings of the IEEE/CVF Conference on Computer Vision and Pattern Recognition (CVPR)},
  pages={56--66},
  year={2024}
}

@article{shapee,
  title={Shap-e: Generating conditional 3d implicit functions},
  author={Jun, Heewoo and Nichol, Alex},
  journal={arXiv preprint arXiv:2305.02463},
  year={2023}
}

@article{shape2vecset,
  title={3dshape2vecset: A 3d shape representation for neural fields and generative diffusion models},
  author={Zhang, Biao and Tang, Jiapeng and Niessner, Matthias and Wonka, Peter},
  journal={ACM Transactions On Graphics (TOG)},
  volume={42},
  number={4},
  pages={1--16},
  year={2023},
  publisher={ACM New York, NY, USA}
}

@article{shapesplat,
  title={Shapesplat: A large-scale dataset of gaussian splats and their self-supervised pretraining},
  author={Ma, Qi and Li, Yue and Ren, Bin and Sebe, Nicu and Konukoglu, Ender and Gevers, Theo and Van Gool, Luc and Paudel, Danda Pani},
  journal={arXiv preprint arXiv:2408.10906},
  year={2024}
}

@inproceedings{perceptual,
  title={The unreasonable effectiveness of deep features as a perceptual metric},
  author={Zhang, Richard and Isola, Phillip and Efros, Alexei A and Shechtman, Eli and Wang, Oliver},
  booktitle={Proceedings of the IEEE conference on computer vision and pattern recognition},
  pages={586--595},
  year={2018}
}

@inproceedings{clip,
  title={Learning transferable visual models from natural language supervision},
  author={Radford, Alec and Kim, Jong Wook and Hallacy, Chris and Ramesh, Aditya and Goh, Gabriel and Agarwal, Sandhini and Sastry, Girish and Askell, Amanda and Mishkin, Pamela and Clark, Jack and others},
  booktitle={International Conference on Machine Learning (ICML)},
  pages={8748--8763},
  year={2021},
  organization={PmLR}
}

@inproceedings{omniobject3d,
  title={Omniobject3d: Large-vocabulary 3d object dataset for realistic perception, reconstruction and generation},
  author={Wu, Tong and Zhang, Jiarui and Fu, Xiao and Wang, Yuxin and Ren, Jiawei and Pan, Liang and Wu, Wayne and Yang, Lei and Wang, Jiaqi and Qian, Chen and others},
  booktitle={Proceedings of the IEEE/CVF Conference on Computer Vision and Pattern Recognition},
  pages={803--814},
  year={2023}
}

@inproceedings{scenesplat,
  title={Scenesplat: Gaussian splatting-based scene understanding with vision-language pretraining},
  author={Li, Yue and Ma, Qi and Yang, Runyi and Li, Huapeng and Ma, Mengjiao and Ren, Bin and Popovic, Nikola and Sebe, Nicu and Konukoglu, Ender and Gevers, Theo and others},
  booktitle={Proceedings of the IEEE/CVF International Conference on Computer Vision},
  pages={4961--4972},
  year={2025}
}

@article{lightgaussian,
  title={Lightgaussian: Unbounded 3d gaussian compression with 15x reduction and 200+ fps},
  author={Fan, Zhiwen and Wang, Kevin and Wen, Kairun and Zhu, Zehao and Xu, Dejia and Wang, Zhangyang and others},
  journal={Advances in neural information processing systems},
  volume={37},
  pages={140138--140158},
  year={2025}
}

@inproceedings{objaverse,
  title={Objaverse: A universe of annotated 3d objects},
  author={Deitke, Matt and Schwenk, Dustin and Salvador, Jordi and Weihs, Luca and Michel, Oscar and VanderBilt, Eli and Schmidt, Ludwig and Ehsani, Kiana and Kembhavi, Aniruddha and Farhadi, Ali},
  booktitle={Proceedings of the IEEE/CVF conference on computer vision and pattern recognition},
  pages={13142--13153},
  year={2023}
}

@inproceedings{structuredistance,
  title={Splicing vit features for semantic appearance transfer},
  author={Tumanyan, Narek and Bar-Tal, Omer and Bagon, Shai and Dekel, Tali},
  booktitle={Proceedings of the IEEE/CVF Conference on Computer Vision and Pattern Recognition (CVPR)},
  pages={10748--10757},
  year={2022}
}

@inproceedings{3dfront1,
  title={3d-front: 3d furnished rooms with layouts and semantics},
  author={Fu, Huan and Cai, Bowen and Gao, Lin and Zhang, Ling-Xiao and Wang, Jiaming and Li, Cao and Zeng, Qixun and Sun, Chengyue and Jia, Rongfei and Zhao, Binqiang and others},
  booktitle={Proceedings of the IEEE/CVF International Conference on Computer Vision (ICCV)},
  pages={10933--10942},
  year={2021}
}

@article{gpt,
  title={Gpt-4o system card},
  author={Hurst, Aaron and Lerer, Adam and Goucher, Adam P and Perelman, Adam and Ramesh, Aditya and Clark, Aidan and Ostrow, AJ and Welihinda, Akila and Hayes, Alan and Radford, Alec and others},
  journal={arXiv preprint arXiv:2410.21276},
  year={2024}
}

@Manual{blender,
   title = {Blender - a 3D modelling and rendering package},
   author = {Blender Online Community},
   organization = {Blender Foundation},
   address = {Stichting Blender Foundation, Amsterdam},
   year = {2018},
   url = {http://www.blender.org},
 }

@inproceedings{mvsplat,
  title={Mvsplat: Efficient 3d gaussian splatting from sparse multi-view images},
  author={Chen, Yuedong and Xu, Haofei and Zheng, Chuanxia and Zhuang, Bohan and Pollefeys, Marc and Geiger, Andreas and Cham, Tat-Jen and Cai, Jianfei},
  booktitle={European Conference on Computer Vision},
  pages={370--386},
  year={2024},
  organization={Springer}
}

@inproceedings{dmd1,
  title={One-step diffusion with distribution matching distillation},
  author={Yin, Tianwei and Gharbi, Micha{\"e}l and Zhang, Richard and Shechtman, Eli and Durand, Fredo and Freeman, William T and Park, Taesung},
  booktitle={Proceedings of the IEEE/CVF Conference on Computer Cision and Pattern Recognition (CVPR)},
  pages={6613--6623},
  year={2024}
}

@article{dmd2,
  title={Improved distribution matching distillation for fast image synthesis},
  author={Yin, Tianwei and Gharbi, Micha{\"e}l and Park, Taesung and Zhang, Richard and Shechtman, Eli and Durand, Fredo and Freeman, Bill},
  journal={Advances in Neural Information Processing Systems (NeurIPS)},
  volume={37},
  pages={47455--47487},
  year={2024}
}

@misc{unityassetstore,
  title        = {Unity Asset Store},
  author       = {{Unity Technologies}},
  howpublished = {\url{https://assetstore.unity.com/}},
  note         = {Accessed: 2026-03-05},
  year         = {2026}
}

@misc{sketchup3dwarehouse,
  title        = {3D Warehouse},
  author       = {{SketchUp}},
  howpublished = {\url{https://3dwarehouse.sketchup.com/}},
  note         = {Accessed: 2026-03-05},
  year         = {2026}
}

\clearpage
\appendix

\begin{center}
    {\LARGE\bfseries Supplementary Material}
\end{center}
\vspace{1em}

\section{Additional Dataset Details}

In this supplementary material, we first present details of the dataset (\cref{datasetdetails}) and implementation (\cref{implementationdetails}). Details of the user study are provided in \cref{userstudy}. Additional experimental results are shown in \cref{experimentalresults}, followed by further ablation studies in \cref{additonalablationstudies}. Lastly, related work (\cref{shapeditor}) and failure cases of our framework (\cref{failurecases}) are further discussed. \textbf{We encourage readers to watch the accompanying demo video showcasing our interactive scene editing performance.}

\section{Additional Dataset Details}
\label{datasetdetails}

In this section, we provide details of the curated dataset used for training, as well as the evaluation benchmark. Edit prompts used for evaluation and for training the single-prompt and multi-prompt models are listed in \cref{tab:editprompt}. 

\noindent\textbf{Training Dataset.}  For training our framework, we employ the TRELLIS-500K \cite{trellis} split, a curated collection of 500K 3D assets primarily from Objaverse (XL) \cite{objaverse}, filtered based on aesthetic scores to support 3D generation tasks. After filtering, the Objaverse portion includes 168K objects with text captions generated using GPT-4o \cite{gpt}, from which we sample 140K assets. Each asset is rendered from 72 viewpoints at a resolution of $400\times400$ in Blender \cite{blender}, and converted into 3D Gaussian splatting representations consisting of 50K Gaussians using LightGaussian \cite{lightgaussian}. To initialize the Gaussian centroids, we uniformly sample a point cloud of 5K points from the 3D asset surfaces. Assets with a PSNR below 30 dB are excluded, resulting in a final dataset of approximately 123K high-quality assets, with 3K reserved for testing and 120K used for training. For latent diffusion pre-training (Stage 2), we use the GPT-generated prompts included in the dataset split. For 3D appearance stylization in latent space (Stage 3), we use a curated style catalog of edit prompts (see \cref{tab:editprompt}).

We will release this refined dataset, which includes multi-view renderings and corresponding 3D Gaussian splatting representations for approximately 123K high-quality assets, to support future research in 3D learning. 

\noindent\textbf{Evaluation Benchmark.}  We sample 30 assets from our unseen test split and apply 10 editing instructions, using each instruction on 3 assets, to compare against test-time optimization methods. Since the training code for Shap-Editor \cite{shapeditor} is not publicly available, we evaluate it separately using its four released pre-trained models each corresponding to a different edit instruction. Specifically, we apply each model to 500 test assets sampled from our test split, resulting in 2000 evaluation pairs in total. 

For the qualitative results of our cross-data generalization experiment, we use the OmniObject3D dataset \cite{omniobject3d}, which comprises 6K objects spanning 190 categories. We render each object from 100 random views at a resolution of $400 \times 400$ in Blender \cite{blender}, and generate the corresponding 3D Gaussian splatting representations using LightGaussian \cite{lightgaussian}. For evaluating generalization to complex real-world scenes, we use the SceneSplat-7K dataset~\cite{scenesplat}, which contains 7916 indoor 3D Gaussian splatting scenes reconstructed from several large-scale RGB-D datasets. The dataset comprises over 11.27 billion 3D Gaussians and includes per-Gaussian annotations. In our experiments, we use scenes from the ScanNet++V2 split and directly operate on the released 3DGS reconstructions.

\begin{table}[h]
  \caption{Edit prompts used in the evaluation benchmark and for training the single-prompt and multi-prompt models.}
  \label{tab:editprompt}
  \centering
  \begin{tabularx}{\linewidth}{>{\RaggedRight\arraybackslash}p{0.20\linewidth} X}
    \toprule
    Evaluation Benchmark &
    \fontsize{8}{9}\selectfont
    \textit{“Make its colours look like rainbow”, 
    “Make it Starry Night Van Gogh painting style”, 
    “Make it wooden”, “Make it in cyberpunk style”, 
    “Make it pop art neon duotone”, “Make it marble”, 
    “Make it Barbie style”, “Make it golden”, 
    “Make it botanical themed”, “Make it Halloween themed”} \\
    \midrule
    Training (single-prompt) &
    \fontsize{8}{9}\selectfont
    Each prompt in the evaluation benchmark is used to train a separate model. \\
    \midrule
    Training (multi-prompt) &
    \fontsize{8}{9}\selectfont
    \textit{“Make its colours look like rainbow”, 
    “Make it Starry Night Van Gogh painting style”, 
    “Make it wooden”, “Make it in cyberpunk style”, 
    “Make it pop art neon duotone”, “Make it marble”, 
    “Make it Barbie style”, “Make it golden”, 
    “Make it botanical themed”, “Make it Halloween themed”, 
    “Make it Hello Kitty themed”, “Make it look like made of steel”, 
    “Make it The Scream Edvard Munch painting”, 
    “Make it in sunset palette”, 
    “Make it Pirates of the Caribbean style”, 
    “Make it look like a statue”, “Make it Garfield themed”, 
    “Make it look like made of bronze”, “Make it Dalmatian style”, 
    "Make it gothic style", "Make it ocean palette", 
    "Make it graffiti street art style", “Make it watercolor painting style”,
    "Turn it into red",
    “Turn it into orange”, "Turn it into pink", "Turn it into blue", 
    "Turn it into yellow", "Turn it into green", "Turn it into black"
    } \\
    \bottomrule
  \end{tabularx}
\end{table}

\section{Additional Implementation Details}
\label{implementationdetails}

\MethodName{} is implemented in PyTorch \cite{pytorch}, and all trainings are performed on 4 NVIDIA H100 GPUs.

\subsection{Architecture Details}

\textbf{Gaussian VAE.} Our Gaussian VAE is built on a grouped-token transformer encoder-decoder backbone following \cite{shapesplat}. In contrast to the original masked autoencoding setup, we reformulate the model as an unmasked VAE. Concretely, we replace masked reconstruction with posterior parameter prediction, reparameterized latent sampling, and a VAE training objective, while scaling the latent representation to $1024 \times 1024$ to match the latent dimensionality required by the subsequent pre-training and editing stages. While GaussianMAE operates on 1024 Gaussians with a latent representation of size $64 \times 384$, such a sparse input is insufficient for our high-dimensional latent modeling. We therefore use 16384 input Gaussians, randomly sub-sampled from each 3DGS asset containing 50K Gaussians. The encoder has 12 transformer layers with 6 attention heads, and the decoder has 4 layers with 6 heads; we use a drop path rate of 0.1. Beyond this backbone, our VAE introduces separate geometry and appearance branches together with a cross-branch feature sharing module for controlled interaction between the two latent spaces. This module consists of a symmetric multi-head cross-attention block with 4 heads, followed by learnable residual scaling and layer normalization, producing fused tokens for decoding. The decoded Gaussians  are rendered with a differentiable Gaussian rasterizer.

\noindent\textbf{Denoiser.} We use a transformer-based text-conditioned denoiser instantiated with the architecture of Shap-E \cite{shapee}. We initialize it from the released pretrained weights to improve training stability and reproducibility, while keeping the overall framework backbone-agnostic. The denoiser operates on latents of shape $1024 \times 1024$. For text conditioning, we prepend a single token containing the CLIP \cite{clip} text embedding. 

\noindent\textbf{Rasterizer.} We employ the parallelizable rasterizer introduced in MVSplat \cite{mvsplat} in our framework. During training, images are rendered at a resolution of $200\times200$ using the camera parameters of the input views provided in the dataset.

\noindent\textbf{2D Image Editor and Auxiliary Denoiser.} A frozen pre-trained InstructPix2Pix (IP2P) \cite{ip2p} model is used as the 2D teacher editor in the latent editing stage. Rendered multi-view images are tiled into a grid and encoded with the frozen Stable Diffusion VAE \cite{stablediffusion} to obtain latents in the 2D diffusion space. IP2P operates on these latents under CLIP \cite{clip} text conditioning. We additionally train an auxiliary denoiser $\epsilon_{\phi}$ with the same conditional U-Net architecture as the IP2P denoiser, initialized from the teacher weights. Although both networks share the same initialization, only the teacher remains frozen, while $\epsilon_{\phi}$ remains trainable.

\subsection{Training Details}

\textbf{Stage 1 - Latent Space Learning.} We train the Gaussian VAE using the AdamW optimizer with an initial learning rate of $1\text{e}^{-3}$ and a weight decay of $5\text{e}^{-2}$. A cosine learning rate scheduler (CosLR) is employed over 200 epochs with a 10-epoch warm-up phase. The model operates on 16384 input Gaussians, grouped into $p = 1024$ clusters of size $k = 32$ using a soft KNN strategy based on Gaussian centroids.  Weight of the fused tokens during injection is set to $\alpha = 0.5$. Training is performed with a batch size of 92. The render loss $\mathcal{L}_{render}$ is computed over $V=6$ random views of the images from which the input Gaussians are optimized. Loss weights $ \left\{ \lambda_{1}, \lambda_{2}, \lambda_{3} \right\}$ are set to $ \left\{ 0.01, 1.0, 1e^{-4} \right\}$. Weight of $\mathcal{L}_{LPIPS}$ is set to $\tau = 0.4$. $\lambda_{4}$ follows a beta-annealing schedule defined as $\lambda_{4} = \min\left(2.0 \cdot \frac{\text{epoch}}{100},\ 1.0\right)$, which gradually increases the $\mathcal{L}_{KL}$ weight to stabilize latent space learning. During training of Stage 2, parameters of the Gaussian VAE are kept fixed and are not updated. In Stage 3, the encoder remains frozen while the decoder is further trained.

\noindent\textbf{Stage 2 - Latent Diffusion Pre-training.} For the latent diffusion model pre-training, we use a lower initial learning rate of $1\text{e}^{-5}$ and the same weight decay of $5\text{e}^{-2}$, along with a cosine annealing learning rate scheduler (CosAnLR) over 200 epochs. Batch size of 120 is employed. The diffusion process is applied only to the appearance latent, while the geometry latent is directly passed to the geometry decoder. To support classifier-free guidance, conditioning is zeroed out with a 0.1 probability during training. The full training objective is:

\begin{equation}
    \mathcal{L} = \lambda_{1}\mathcal{L}_{param} + \lambda_{2}\mathcal{L}_{render} + \lambda_{5}\mathcal{L}_{diff}, 
\end{equation}
where $ \left\{ \lambda_{1}, \lambda_{2}, \lambda_{5} \right\}$ are set to $ \left\{ 0.01, 1.0, 1.0\right\}$, and $\mathcal{L}_{diff}$ is the diffusion loss from Equation 9 of the main paper.

\noindent\textbf{Stage 3 - 3D Appearance Stylization in Latent Space.} Following the DMD objective \cite{dmd1,dmd2}, we adapt distribution matching to our latent editing setting. Compared with the original DMD formulation, our setting differs in three aspects: (i) distribution matching is performed in the Stable Diffusion latent space, (ii) the student representation is obtained by encoding tiled multi-view renderings to encourage 3D consistency, and (iii) training is stabilized with auxiliary teacher-image anchoring and color-variance regularization.

\noindent\textit{\textbf{DMD Formulation.}} We perform distribution matching in the Stable Diffusion latent space as both the frozen IP2P teacher and the auxiliary denoiser operate naturally in this space. For each training iteration, the current 3DGS prediction of \MethodName{} is rendered from $V$ viewpoints and the resulting images are tiled into a multi-view grid. The grid is encoded using the frozen Stable Diffusion VAE to obtain a latent representation $y$. A timestep $t$ is sampled from the diffusion schedule and Gaussian noise is added to obtain the noised latent $y_t$. 
The frozen IP2P denoiser serves as the teacher and predicts the noise $\epsilon_{\mathrm{real}}(y_t,t)$ under the edit text and image conditioning. An auxiliary denoiser $\epsilon_{\phi}$, with the same conditional U-Net architecture, predicts $\epsilon_{\mathrm{fake}}(y_t,t)$ under the same conditioning for the DMD update. These two noise predictions are converted into clean-latent estimates, $\hat{x}_{0,\mathrm{real}}$ and $\hat{x}_{0,\mathrm{fake}}$, and the DMD update for the student latent is applied in $\hat{x}_0$-space, as described in Equation 10 of the main paper. Separately, the auxiliary denoiser is optimized on detached, independently noised student latents with an SNR-weighted diffusion noise prediction objective, allowing it to track the evolving student distribution without backpropagating through the student update. Since this auxiliary denoising loss is optimized independently, it is not included in the latent editing objective in Equation 11.

\noindent\textit{\textbf{Training Stabilization.}} Using the same tiled multi-view grid described above, we additionally obtain teacher-image supervision from the frozen IP2P model to stabilize training. Let $\mathbf{V}^{\mathrm{grid}}_{0}$ denote the grid of rendered source views. Conditioned on the edit text embedding $\mathcal{C}^{e}$ and the source image grid, the frozen IP2P teacher produces an edited grid:
\vspace{-0.3em}
\begin{equation}
    \tilde{\mathbf{V}}^{\mathrm{grid}} = E_{\theta}\!\left(\mathbf{V}^{\mathrm{grid}}_{0}; \mathcal{C}^{e}\right),
\end{equation}
\noindent where $E_{\theta}$ denotes the frozen IP2P editing process and $\tilde{\mathbf{V}}^{\mathrm{grid}}$ is the resulting edited multi-view grid. Similar to the computation of $\mathcal{L}_{rgb}$, the teacher-image anchoring loss $\mathcal{L}_{MV\text{-}L1}$ is computed over the $V$ edited views and their corresponding rendered counterparts by \MethodName{}. This auxiliary teacher-image anchoring term improves optimization stability, especially in the early stages. To further prevent degenerate single-color solutions, we apply a color-variance regularizer on the renderings from \MethodName{}:
\vspace{-0.3em}
\begin{equation}
    \mathcal{L}_{\mathrm{reg}}
    =
    \frac{1}{BV}\sum_{b=1}^{B}\sum_{v=1}^{V}
    \max\!\left(0, \tau - \bar{\sigma}_{b,v}\right),
\end{equation}
\noindent where $\bar{\sigma}_{b,v}$ denotes the foreground-weighted mean RGB standard deviation of the student rendering for view $v$ in sample $b$, and $\tau$ is a target standard deviation. 

\noindent\textit{\textbf{Optimization Details.}} Similarly, this stage is trained using a lower initial learning rate of $1\mathrm{e}{-5}$ and the same weight decay of $5\mathrm{e}{-2}$, using a CosAnLR scheduler over 10 epochs. A batch size of 90 is used. For the DMD update, the diffusion timestep $t$ is sampled from the range $[0.02,\ 0.70]$.  For InstructPix2Pix~\cite{ip2p}, we use classifier-free guidance with text guidance scale $s_T \in [5.5,\ 9.5]$ and image guidance scale $s_I = 2.0$, where $s_T$ is randomly sampled for each batch. Training uses $V=6$ rendered views arranged into a tiled multi-view grid, and this stage is optimized with the latent editing objective defined in the main paper. We set the teacher-image anchoring weight to $\lambda_{MV\text{-}L1}=0.1$ and the color-variance regularization weight to $\lambda_{\mathrm{reg}}=0.1$. The teacher-image anchoring loss is applied sparsely at fixed training intervals for stabilization, while the color-variance regularizer is applied throughout training.

\subsection{Evaluation Details}

To evaluate how well the 3D edits align with the edit prompts, we report CLIP \cite{clip} similarity ($\mathrm{CLIP}_{\mathrm{sim}}$) and directional similarity ($\mathrm{CLIP}_{\mathrm{dir}}$) scores. Following Shap-Editor \cite{shapeditor}, we also report Structure Distance (Structure Dist.) \cite{structuredistance} metric to quantify structural consistency between input and edited assets. To compute the scores, outputs from our framework and the baselines are rendered at a resolution of $200 \times 200$ from 8 viewpoints using the camera parameters provided in the dataset.

\section{Details of the User Study}
\label{userstudy}

\newcolumntype{R}[1]{>{\RaggedRight\arraybackslash}p{#1}}
\newcolumntype{F}[1]{>{\RaggedRight\arraybackslash}p{#1}}
\begin{table}[h]
  \caption{\textbf{The criteria used in the user study.} The definitions listed here are also provided to participants.
  }
  \label{tab:suppcriteria}
  \centering
  \begin{tabular}{R{0.24\textwidth}F{0.70\textwidth}}
    \toprule
    \textbf{Criterion} &  \\
    \midrule 
    Text Alignment & Which stylized result follows the prompt most closely? \textit{Consider whether the style change is too subtle (almost no visible change), or too drastic (over-saturated).} \\ 
    \hline 
    Structural Consistency & Which edited result best preserves the original asset's 3D structure while applying only the requested appearance stylization? \textit{The object’s shape should not change after stylization (e.g., no changes to the object’s physical structure; for characters, the face, identity and features, must remain unchanged).} \\
    \hline 
    Visual Quality & Which edited result looks best overall?  \\
  \bottomrule
  \end{tabular}
\end{table}

We conducted a user study with 50 participants to assess perceptual preferences between \MethodName{} and optimization-based methods IN2N \cite{in2n}, IGS2GS \cite{igs2gs}, GaussCtrl \cite{gaussctrl} and GaussianEditor \cite{gaussianeditor}. Shap-Editor~\cite{shapeditor} was excluded because publicly available models do not cover the study's edit range. For each of the 8 tasks, participants viewed outputs from \MethodName{} and the baselines from multiple viewpoints, alongside the edit prompt and the input 3D asset. For each criterion, participants were requested to select the output with the best (i) text-alignment, (ii) structural consistency, and (iii) visual quality, with detailed definitions provided in \cref{tab:suppcriteria}. We further asked the acceptable per-asset wait time conditioned on their preferred results.

The results in Tab. 2 of the main paper show that \MethodName{} was consistently preferred across all three criteria, confirming its strength in text-to-3D appearance stylization compared with test-time optimization methods. For latency, 69.6\% favored real-time ($<1$ s); the remaining choices were 2–10 mins (17.4\%), 10–20 mins (13.0\%), and 20–60 mins (0\%), indicating a clear preference for fast editing.

\section{Additional Experimental Results}
\label{experimentalresults}

\subsection{Effect of Style Catalog Size}
\label{multiprompt}

\begin{figure}[h!]
  \centering
   \includegraphics[height=4.5cm]{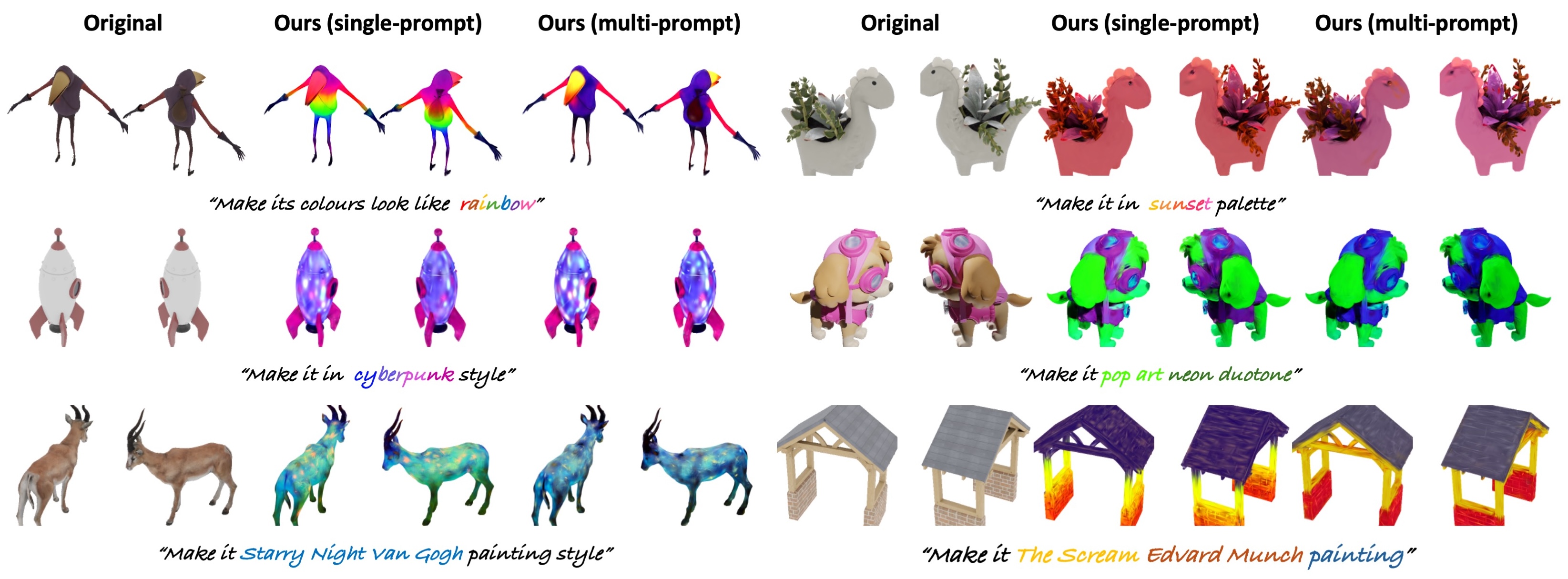}
   \caption{\textbf{Visual results of expanding the style catalog.} \MethodName{} maintains high-quality 3D stylization and geometry preservation as the number of training prompts increases.}
   \label{fig:suppmultiprompt}
\end{figure}

In the main paper, we show quantitatively that \MethodName{} scales from 1 to 30 training prompts while maintaining strong stylization performance, supporting its use with larger curated style catalogs. Here, we provide additional visual comparisons to complement this analysis. The full set of edit instructions is given in \cref{tab:editprompt}. Visual comparisons between the single-prompt and 30-prompt variants of \MethodName{} in \cref{fig:suppmultiprompt} show that the style catalog can be expanded while maintaining visual quality and geometric fidelity.

\subsection{More Visual Results}
\label{morevisualresults}

\begin{figure}[h!]
  \centering
   \includegraphics[height=10.8cm]{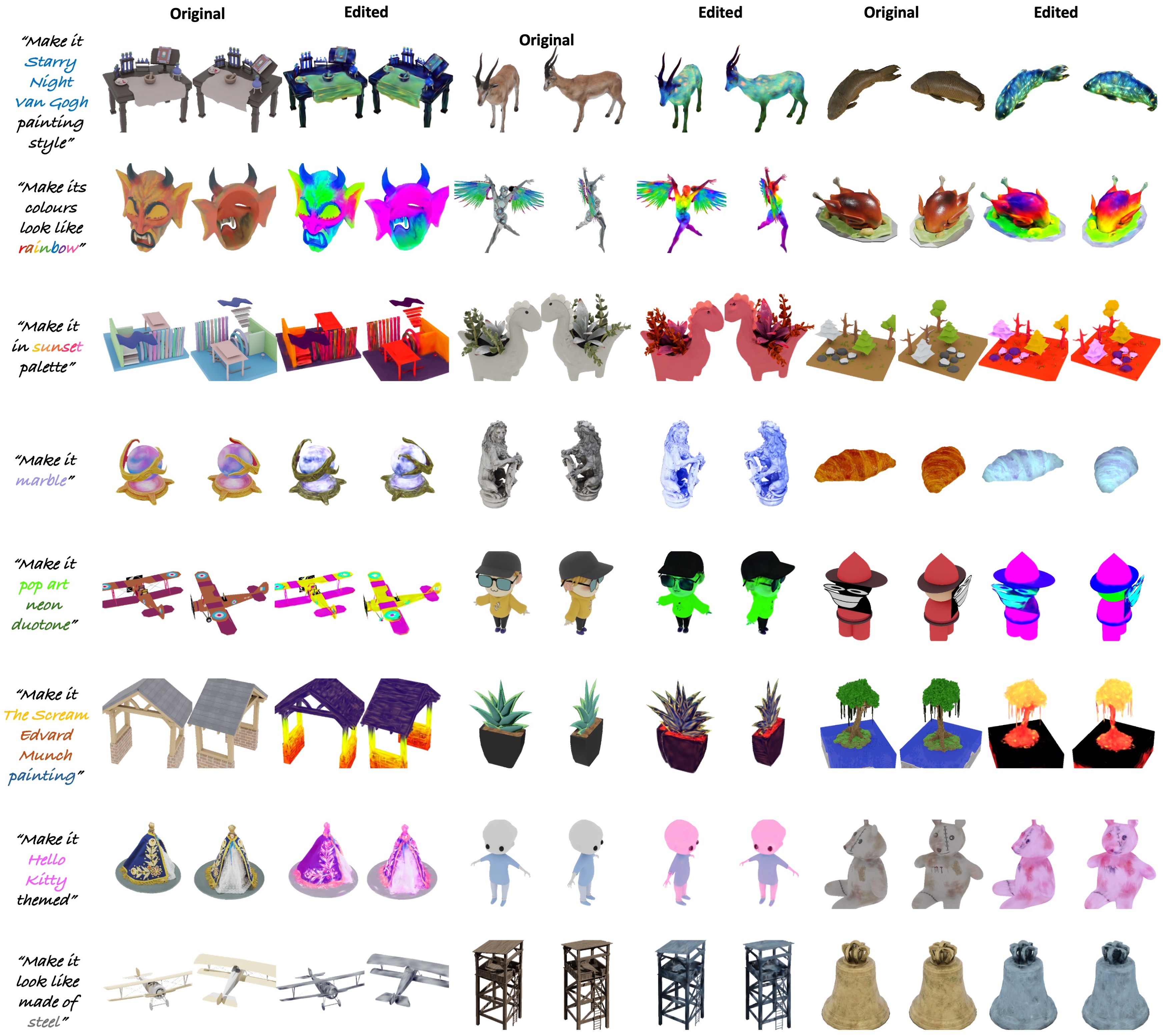}
   \caption{\textbf{Additional visual results of our method.} \MethodName{} delivers instant, geometry-preserving, and multi-view-consistent 3D stylizations in a single feed-forward pass.}
   \label{fig:suppqual1}
\end{figure}

\begin{figure}[h!]
  \centering
   \includegraphics[height=9.5cm]{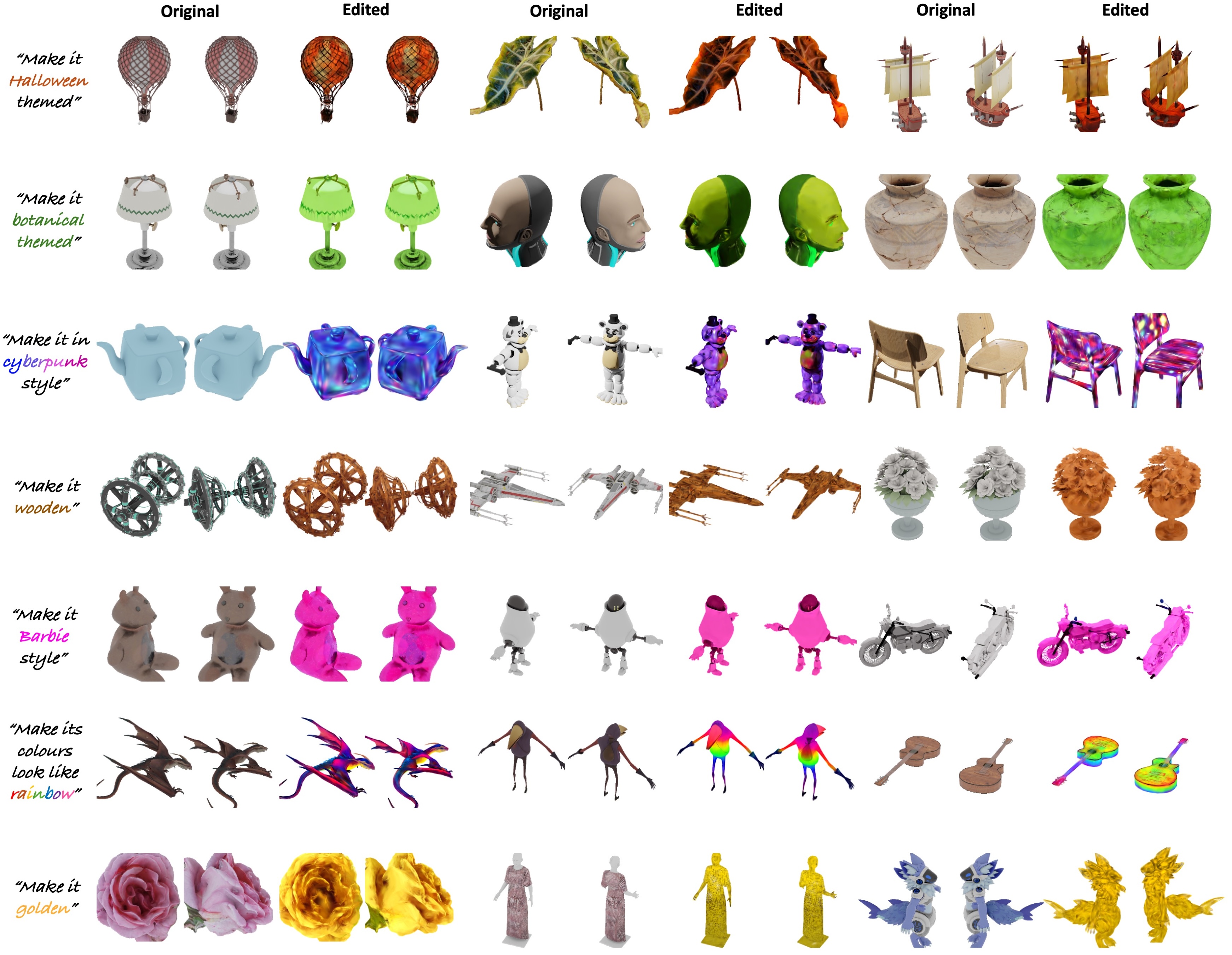}
   \caption{\textbf{Additional visual results of our method.} \MethodName{} delivers instant, geometry-preserving, and multi-view-consistent 3D stylizations in a single feed-forward pass.}
   \label{fig:suppqual2}
\end{figure}

We provide additional visual results of our method in \cref{fig:suppqual1} and \cref{fig:suppqual2}. \MethodName{} consistently delivers high-quality, text-aligned 3D stylizations across a wide range of 3D assets in a single forward pass, eliminating the need for test-time optimization. By decoupling appearance from geometry with controlled cross-branch exchange, it enables targeted edits while preserving geometry, making it well suited to direct stylization of existing 3D assets at scale across diverse styles.

\section{Additional Ablation Studies}
\label{additonalablationstudies}

\begin{figure}[h!]
  \centering
   \includegraphics[height=2.0cm]{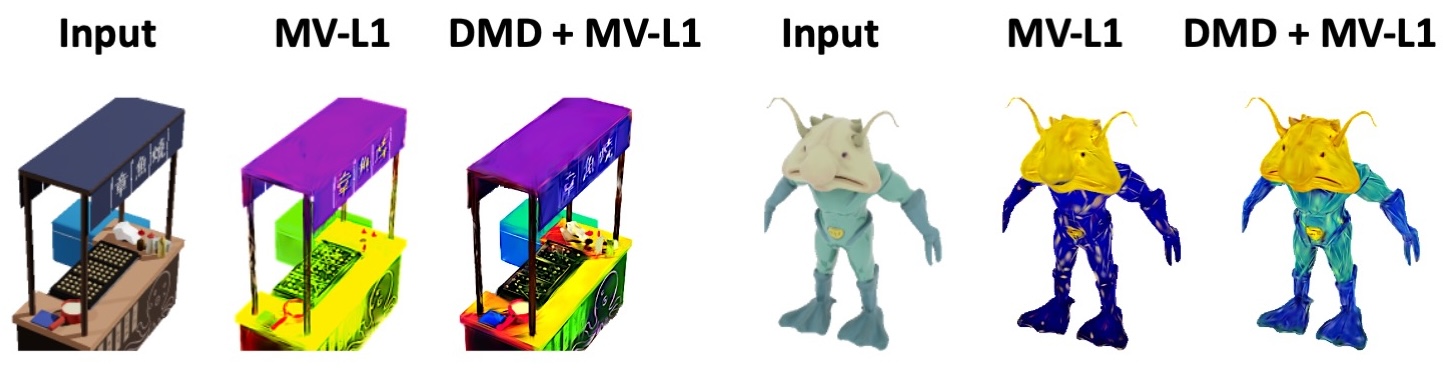}
   \caption{\textbf{Effect of DMD over MV-L1 Supervision.} Compared with the MV-L1 baseline, adding the DMD objective yields sharper, more vivid stylization results.}
   \label{fig:dmd_supp}
\end{figure}

\noindent\textbf{Effect of DMD over MV-L1 Supervision.} We compare a baseline trained with only teacher-image anchoring supervision ($\mathcal{L}_{MV\text{-}L1}$) and the regularization terms, against our full model that additionally includes the DMD objective in \cref{tab:DMD}. The MV-L1 baseline already provides stable supervision from multi-view IP2P edited targets, incorporating DMD improves performance by enforcing distribution matching in latent space, yielding stronger stylization while preserving multi-view consistency. Qualitatively, the DMD variant also produces sharper and more vivid results (\cref{fig:dmd_supp}), indicating that DMD provides complementary supervision beyond teacher-image anchoring alone.

\begin{table}[h]
\caption{\textbf{Effect of DMD over MV-L1 Supervision.} DMD improves metrics over MV-L1 supervision.}
\centering
    \resizebox{0.7\linewidth}{!}{
    \begin{tabular}{l|lll}
    \toprule
    \multicolumn{1}{c|}{ }  &  \multicolumn{1}{c}{$\mathrm{CLIP}_{\mathrm{sim}}\!\uparrow$} & \multicolumn{1}{c}{$\mathrm{CLIP}_{\mathrm{dir}}\!\uparrow$} & \multicolumn{1}{c}{Structure Dist. $\downarrow$}     \\
    \hline \hline   
    MV-L1 &  0.251 & 0.210 & 0.0064   \\
    DMD + MV-L1 (Ours) & \textbf{0.260}  & \textbf{0.223} & \textbf{0.0050}   \\
    \bottomrule
    \end{tabular}}
\label{tab:DMD}
\end{table}

\noindent\textbf{Effect of Guidance Scale.} \MethodName{} distills from InstructPix2Pix~\cite{ip2p} in Stage 3, using classifier-free guidance with text scale $s_T\!\in\![5.5,9.5]$ and image scale $s_I\!=\!2.0$. We compare our main model against higher text-scale variants: (2) $s_T\!\in\![9.5,12.5]$ and (3) $s_T\!\in\![12.5,15.5]$. In all settings, $s_T$ is sampled uniformly per batch within the specified interval. (1) $s_T\!\in\![5.5,9.5]$ denotes the main model used throughout the paper. As shown in \cref{tab:textguidancescale}, higher text guidance scales yield improved CLIP metrics with minimal loss in structural consistency.

\begin{table}
  \caption{\textbf{Effect of guidance scale.} Higher text guidance scales yield improved CLIP metrics with minimal loss in structural consistency. }
  \label{tab:textguidancescale}
  \centering
  \begin{tabular}{@{}lccc@{}}
    \toprule
      & $\mathrm{CLIP}_{\mathrm{sim}}\!\uparrow$ & $\mathrm{CLIP}_{\mathrm{dir}}\!\uparrow$ & \makecell{Structure \\ Dist. $\downarrow$} \\
    \midrule
    (1) $s_T\!\in\![5.5,9.5]$  & 0.260 & 0.223 & 0.0050 \\
    (2) $s_T\!\in\![9.5,12.5]$   & 0.263 & 0.226 & 0.0089 \\
    (3) $s_T\!\in\![12.5,15.5]$ & 0.268 & 0.232 & 0.0172 \\
    \bottomrule
  \end{tabular}
\end{table}

\begin{figure}[h]
  \centering
  \begin{minipage}{\linewidth}
    \centering
    \includegraphics[height=4.0cm]{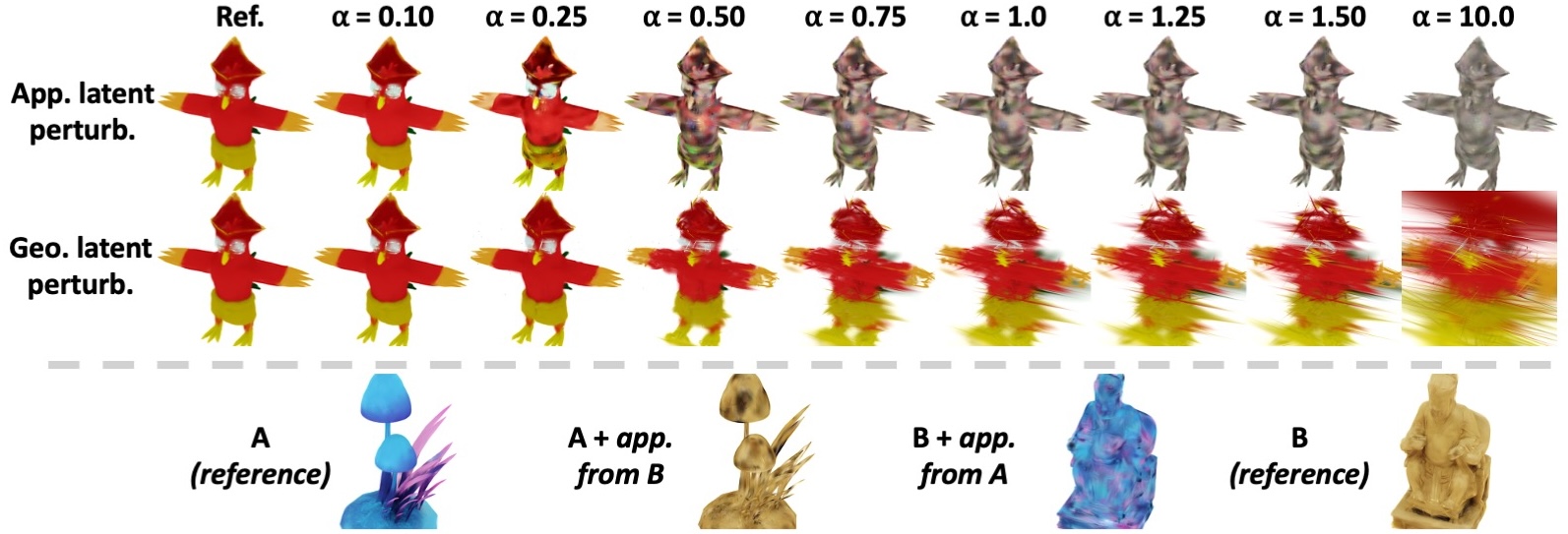}
    \caption{\textbf{Latent perturbation and appearance swapping.} Perturbations affect the corresponding appearance or geometry factor, while appearance swaps transfer appearance while preserving geometry.}
    \label{fig:suppperturbfig}
  \end{minipage}
\end{figure}

\noindent\textbf{Effect of Perturbation on the Latents.} Beyond the latent cosine separation by $\mathcal{L}_{LD}$, we further evaluate disentanglement at the model output. As shown in Fig.~\ref{fig:suppperturbfig}, perturbing the appearance latent $\mathbf{z}_c$ progressively changes appearance while preserving geometry, whereas perturbing the geometry latent $\mathbf{z}_g$ changes geometry while preserving appearance, even under large perturbations up to $\alpha=10$. Similarly, swapping appearance latents transfers visual appearance while largely preserving geometry, with a silhouette IoU of 0.985. Together, these results show that the learned factorization is reflected directly in the model behavior rather than arising only from downstream performance.

\section{Further Discussion on Prior Work}
\label{shapeditor}

\MethodName{} departs from prior state-of-the-art methods in several key aspects:

\textbf{1) Feed-forward design:} Our method is a feed-forward 3DGS style editor that operates in the latent space of a 3D Gaussian VAE, fully eliminating test-time optimization. Optimization-based methods often require minutes to tens of minutes per asset, while \MethodName{} achieves $\sim$0.26 second latency, enabling both large-scale production of diverse assets and truly interactive editing, both of which are crucial for practical 3D content creation workflows. Furthermore, \MethodName{} is distilled from a 2D editor with guidance scales sampled from a reasonable range across many samples. As a result, it is far less sensitive to the 2D editor’s guidance scale than optimization-based baselines, which typically demand careful tuning to avoid overly drastic or barely perceptible edits. 

\textbf{2) Disentangled appearance–geometry with controlled sharing:} \MethodName{} learns separate latent spaces for appearance and geometry, coupled with a lightweight feature-sharing module that enables controlled information exchange. This focuses edits on appearance while effectively preserving the underlying geometry. As shown in our ablations, removing disentanglement degrades reconstruction quality, and removing the feature-sharing module harms editing performance. In contrast, Shap-Editor~\cite{shapeditor}, while also avoiding test-time optimization, operates on a shared appearance–geometry representation encoded from NeRFs using a frozen Shap-E~\cite{shapee} autoencoder. As evident from Shap-Editor’s results, this shared representation is ill-suited for global appearance editing; it limits control and often produces extreme smoothness, loss of fine details, and severe structural distortion. Our disentangled formulation is therefore better aligned with geometry-preserving, large-scale stylization.

\textbf{3) Structured latent representation for 3D Gaussians:} As discussed, operating directly on 3D Gaussians is challenging due to their unstructured nature and the uneven distribution of geometry and appearance parameters, which complicates joint optimization. \MethodName{} makes this feasible by grouping Gaussians by spatial proximity and encoding them into disentangled latent spaces that retain global 3D structure. These structured latent representations provide a promising starting point for downstream 3D applications and future research.

\section{Failure Cases}
\label{failurecases}

\begin{figure}[h]
  \centering
   \includegraphics[height=2.05cm]{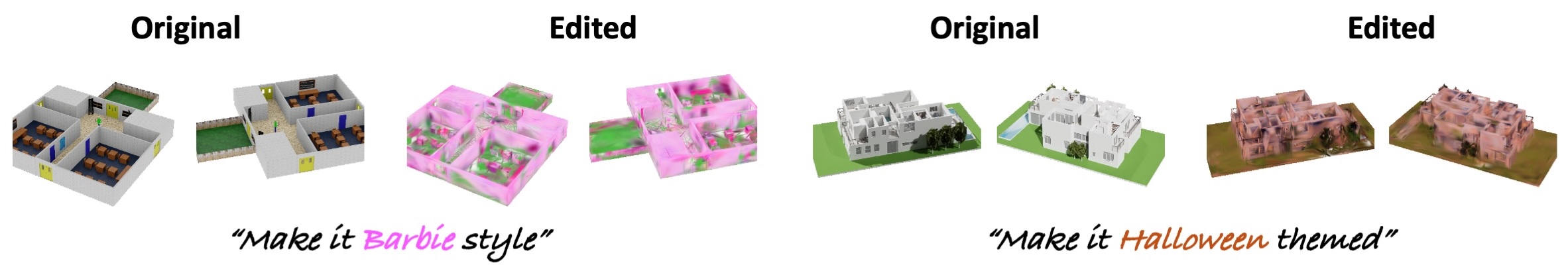}
   \caption{\textbf{Failure cases.} Our model struggles to represent complex scenes with fine-grained details using 16384 Gaussians, leading to lower-quality results. This can be addressed by increasing the number of Gaussians to better capture such details.}
   \label{fig:suppfail}
\end{figure}

The failure cases of our model are shown in Figure~\ref{fig:suppfail}. Since our model represents assets using 16384 Gaussians, it may be insufficient for large complex assets such as full scenes, leading to lower quality reconstructions and blurry edits. These cases can be addressed by increasing the number of Gaussians to better capture fine-grained details in complex scenes.

\end{document}